\documentclass[11pt]{article}

\usepackage[preprint]{acl}

\usepackage{times}
\usepackage{latexsym}

\usepackage[T1]{fontenc}

\usepackage[utf8]{inputenc}

\usepackage{microtype}

\usepackage{inconsolata}

\usepackage{graphicx}
\usepackage{amsmath}
\usepackage{booktabs}
\usepackage{subcaption}
\usepackage{listings}
\usepackage{cuted}
\usepackage{amssymb}
\usepackage{cleveref}
\usepackage{tabularx}
\usepackage{multirow}

\newcommand{\name}{SCMAPR}

\usepackage[most]{tcolorbox} 

\tcbset{
	mybox1/.style={
		colback=black!5!white, 
		colframe=black, 
		colbacktitle=black, 
		coltitle=white, 
		fonttitle=\bfseries, 
		boxsep=5pt, 
		left=5mm, 
		right=5mm, 
		top=5mm, 
		bottom=5mm, 
		arc=3mm, 
		auto outer arc, 
		boxrule=2pt, 
	}
}

\newtcolorbox{myboxwide}[2][]{
  mybox1,             
  enhanced,           
  float*=htbp,        
  width=\textwidth,   
  title={#2},         
  #1                  
}

\title{\name{}: Self-Correcting Multi-Agent Prompt Refinement for Complex-Scenario Text-to-Video Generation}

\author{
 \textbf{Chengyi Yang\textsuperscript{1,2}\footnotemark[1]},
 \textbf{Pengzhen Li\textsuperscript{1}},
 \textbf{Jiayin Qi\textsuperscript{3}},
 \textbf{Aimin Zhou\textsuperscript{2}},
 \textbf{Ji Wu\textsuperscript{4}},
 \textbf{Ji Liu\textsuperscript{1}\footnotemark[2]}
\\
\\
 \textsuperscript{1}HiThink Research,
 \textsuperscript{2}East China Normal University,
 \\
 \textsuperscript{3}Guangzhou University,
 \textsuperscript{4}Tsinghua University
\\
}

\begin{document}
\maketitle

\begingroup
\renewcommand{\thefootnote}{\fnsymbol{footnote}}
\footnotetext[1]{Work done during his internship at HiThink Research under supervision of Ji Liu.}
\footnotetext[2]{Corresponding author: jiliuwork@gmail.com}
\endgroup

\begin{abstract}
Text-to-Video (T2V) generation has benefited from recent advances in diffusion models, yet current systems still struggle under complex scenarios, which are generally exacerbated by the ambiguity and underspecification of text prompts. In this work, we formulate complex-scenario prompt refinement as a stage-wise multi-agent refinement process and propose \name{}, i.e., a scenario-aware and Self-Correcting Multi-Agent Prompt Refinement framework for T2V prompting. \name{} coordinates specialized agents to (i) route each prompt to a taxonomy-grounded scenario for strategy selection, (ii) synthesize scenario-aware rewriting policies and perform policy-conditioned refinement, and (iii) conduct structured semantic verification that triggers conditional revision when semantic fidelity violations are detected. To clarify what constitutes complex scenarios in T2V prompting, provide representative examples, and enable rigorous evaluation under such challenging conditions, we further introduce T2V-Complexity, which is a complex-scenario T2V benchmark consisting exclusively of complex-scenario prompts.
Extensive experiments on 3 existing benchmarks and our T2V-Complexity benchmark demonstrate that \name{} consistently improves text-video alignment and overall generation quality under complex scenarios, achieving up to 2.67\% and 3.28 gains in average score on VBench and EvalCrafter, and up to 0.028 improvement on T2V-CompBench over 3 state-of-the-art baselines. The codes of \name{} are publicly available at \url{https://github.com/HiThink-Research/SCMAPR}.
\end{abstract}

\section{Introduction}
\label{sec:intro}

The rapid advancement of diffusion models \citep{peebles2023scalable,rombach2022high} has revolutionized Artificial Intelligence Generated Content (AIGC), with applications ranging from image and video generation to 3D content creation \citep{DBLP:conf/cvpr/GuoHHBGS25,DBLP:conf/cvpr/HuangGAY0ZLLCS25,DBLP:conf/cvpr/Lin0CXYWXXZC25,DBLP:conf/cvpr/ZhangLFLD25}, speech and audio synthesis \citep{DBLP:conf/nips/LuoYHZ23,DBLP:conf/mm/LiuLL0024,DBLP:journals/taslp/OhLL24}, and controllable editing \citep{DBLP:conf/acl/LeeKYY25,DBLP:conf/cvpr/HeWG25,DBLP:conf/aaai/WangLLZL0C25}, driving new opportunities in entertainment, education, design, and human–computer interaction. As an AIGC modality, Text-to-Video (T2V) generation requires visually realistic frames together with temporal coherence, motion dynamics, and adherence to physical and causal constraints, thereby posing a substantially challenging yet impactful open research problem.

Existing studies~\citep{hao2023optimizing,zhan2024prompt,zhan2024capability} demonstrate that optimizing prompts with Large Language Models (LLMs) leads to high-quality and user-aligned generations in diffusion-based content creation. Such improvements are particularly evident when user inputs are concise and underspecified. 
In practice,  high-quality prompts specify characters and scenes precisely, follow effective expression patterns, and incorporate domain-specific terminology for stylistic control \citep{parsons2022dalle2,witteveen2022investigating,brade2023promptify,zhan2024capability}.

However, not all T2V generation tasks can be materially improved by merely expanding the input text. 
Complex-scenario T2V generation tasks differ from common T2V tasks in the target video scenario. 
For complex-scenario T2V generation tasks, the target video scenario is dominated by one or multiple specific challenging sources for the current T2V diffusion model to realize with temporal coherence. To make this notion explicit, we introduce a ten-category taxonomy of complex scenarios and use the dominant category as a routing signal for policy generation and prompt refinement. Please see formal definitions and representative examples in Appendix~\ref{app-sec:complex_scenario_taxonomy}. 

Meanwhile, existing prompt refinement approaches are generally designed for Text-to-Image (T2I) tasks, which typically involve employing retrieval-augmented generation to expand prompts~\citep{sun2024retrieval}, tailoring prompts based on user preferences~\citep{zhan2024prompt,zhan2024capability}, and enhancing prompts through entity-specific descriptions~\citep{ozaki2025texttiger}. While effective for single-image synthesis, these strategies are insufficient for T2V generation. Unlike T2I, T2V tasks are required to simultaneously guarantee temporal consistency, ensure motion coherence, capture causal dependencies, and comply with physical laws across frames. Although a few prompt refinement approaches exist for T2V, e.g., Retrieval-Augmented Prompt Optimization (RAPO)~\citep{gao2025devil}, they may focus on inter-object relations while overlooking video-specific challenges, e.g., abstract semantics and temporal consistency, in complex scenarios. 
In addition, existing T2V benchmarks are not designed for complex scenarios. They typically do not fully cover complex-scenario categories and exhibit severe category imbalance, which hinders rigorous and systematic evaluation under complex scenarios.

In this paper, we formulate the prompt refinement process for complex-scenario T2V generation as a \emph{stage-wise multi-agent refinement process}. In this process, specialized agents collaboratively identify the dominant source of difficulty via scenario tagging. Then, they plan scenario-appropriate refinement strategies and verify semantic fidelity through structured analysis. Motivated by this perspective, we develop a Self-Correcting Multi-Agent Prompt Refinement framework (\name{}) with a pipeline of five stages, i.e., (1) taxonomy-based scenario routing, (2) scenario-aware policy synthesis, (3) policy-conditioned prompt refinement, (4) structured semantic verification, and (5) conditional revision according to validation results. To clarify complex scenarios in T2V prompts with representative examples, we introduce the benchmark {T2V-Complexity}, consisting exclusively of prompts balanced across complex-scenario categories. We note that multi-agent prompt refinement has been explored for complex T2I generation~\citep{DBLP:conf/cvpr/LiHLYQCWJXZ25}. In contrast, our method targets T2V and introduces verification-driven self-correction rather than pure information extraction.

Overall, \name{} is designed to enable scenario-aware and self-correcting prompt refinement for complex-scenario T2V generation. \name{} exploits structured and verification-driven signals to guide conditional revision, thereby improving semantic fidelity under challenging prompts. Together with the proposed T2V-Complexity benchmark, \name{} supports systematic study and evaluation of complex-scenario T2V prompting. The main contributions are summarized as follows:

\begin{itemize}
\item  We introduce a scenario-aware refinement pipeline for complex-scenario T2V prompt refinement, including taxonomy-grounded routing, prompt-specific policy synthesis, and policy-conditioned rewriting (Stages I-III).

\item  We propose a verification-driven self-correction design that performs atom-level semantic verification and conditional prompt revision to improve  user-intent preservation and semantic fidelity (Stages IV-V).


\item We introduce T2V-Complexity, a taxonomy-grounded benchmark consisting exclusively of complex-scenario prompts and balanced across ten complex-scenario categories, enabling rigorous evaluation of T2V generation under diverse complex scenarios.

\item We conduct extensive experimentation on 3 existing benchmarks and T2V-Complexity, demonstrating that \name{} consistently improves semantic fidelity and overall video generation quality for T2V generation.
\end{itemize}


\section{Related Work}
In this section, we present existing works on T2V generation and prompt refinement approaches.

\subsection{Text-to-Video Generation}

Recent advancements in diffusion transformers and large-scale generative models~\citep{rombach2022high,peebles2023scalable} have significantly promoted T2V generation~\citep{singer2022make,chen2024videocrafter2}.
Prior works have explored a variety of directions, including scalable architectures such as expert transformers and linear-complexity attention modules~\citep{yang2024cogvideox,wang2025lingen}, training-free and plug-and-play inference techniques for improving motion dynamics and spatial fidelity~\citep{bu2025bytheway,zhang2025videoelevator,jagpal2025eidt}, as well as structured captions, instance-aware modeling and Low-Rank Adaptation (LoRA)-based customization for controlling entity appearance and interactions~\citep{feng2025blobgen,fan2025instancecap,huang2025videomage}.
Beyond visual quality, LLM-guided reasoning and external knowledge retrieval have also been introduced to enhance physical plausibility and factual correctness~\citep{xue2025phyt2v,yuan2025identity}. Despite these advances, most existing T2V approaches implicitly assume relatively common scenarios with well-specified and explicit prompts. 
As a result, T2V generation under complex scenarios, including those involving abstract semantics, intricate multi-entity interactions, and long-range temporal dependencies, remains challenging.

\subsection{Prompt Refinement}

Prompt refinement aims to transform user inputs into formulated prompts that align with the preferences and capabilities of diffusion models. 
Early studies focus on inferring user preferences or rewriting patterns to guide prompt refinement. Representative approaches, such as PRIP~\citep{zhan2024prompt} and CAPR~\citep{zhan2024capability}, adapt prompts based on inferred user capabilities or configurable features, but rely heavily on user interaction data, system logs, or large-scale feedback, which limits their applicability in settings without explicit user supervision. More recent approaches incorporate Retrieval-Augmented Generation (RAG) to enrich prompts by retrieving semantically relevant descriptions.
These methods either maintain external prompt repositories~\citep{sun2024retrieval} or construct relation graphs from training data to retrieve semantically related terms~\citep{gao2025devil}, which are subsequently processed and incorporated into the user input to produce an expanded reformulation.
While effective for improving descriptive richness, such relevance-driven retrieval and augmentation strategies primarily expand prompts based on surface similarity, without explicitly reasoning about the underlying sources of difficulty or specifying scenario-specific refinement guidelines.

\begin{figure*}[t]
    \centering 
    \includegraphics[width=\textwidth]{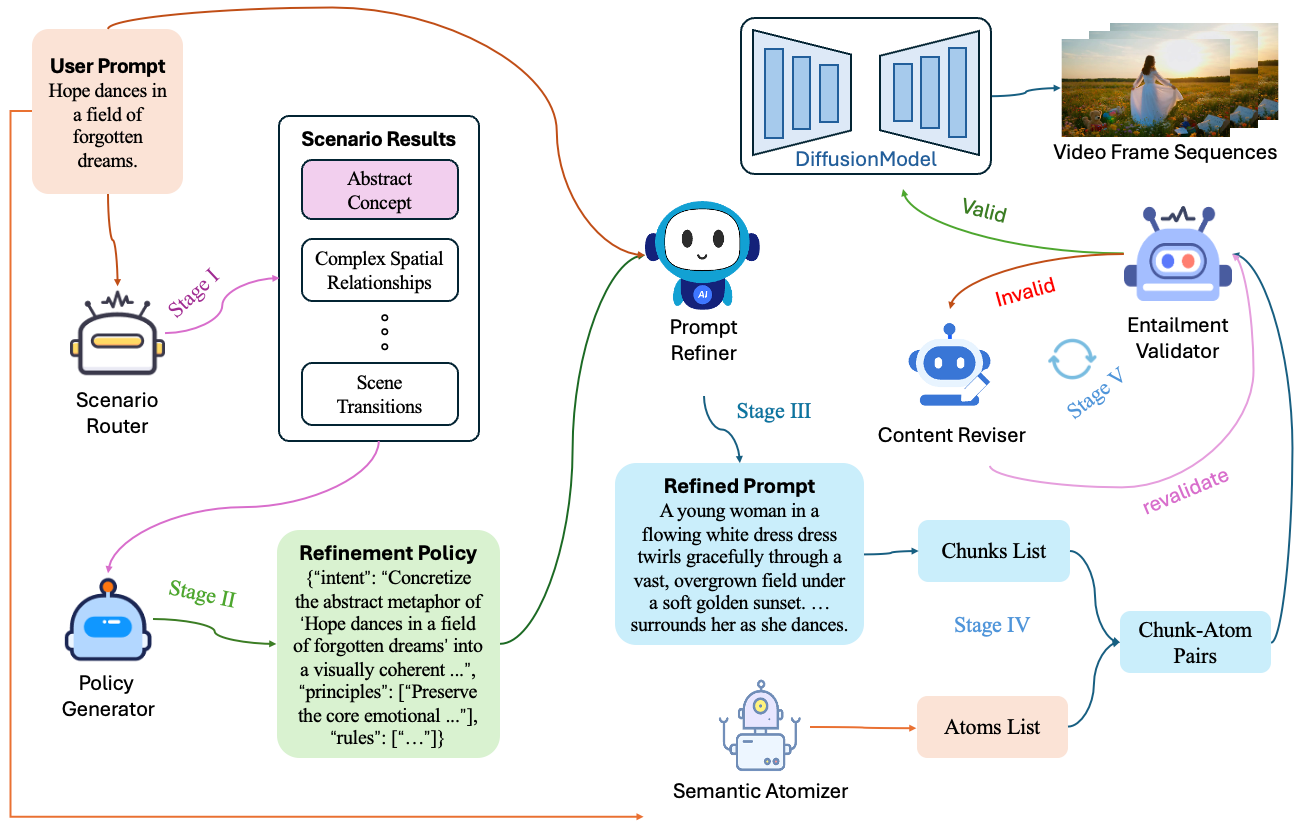}
    \vspace{-7mm}
    \caption[Self-Correcting Multi-Agent Prompt Refinement Framework]{ \textbf{Self-Correcting Multi-Agent Prompt Refinement Framework (\name{}).}
    \name{} organizes prompt refinement as a stage-wise multi-agent collaboration involving six specialized agents. The framework proceeds through five functional stages: (I) \emph{Scenario Routing}, where Scenario Router assigns a scenario tag to the input prompt. (II) \emph{Policy Synthesis}, where a Policy Generator generates a scenario-conditioned rewriting policy. (III) \emph{Policy-Conditioned Refinement}, where a Prompt Refiner rewrites the prompt. (IV) \emph{Semantic Verification}, where Atomizer and Validator collaboratively verify semantic fidelity through atomic extraction and entailment judgment. (V) \emph{Conditional Revision}, where verification feedback conditionally triggers targeted revision, enabling self-correcting refinement.
  }
  \vspace{-5mm}
  \label{fig:refinement-pipeline} 
\end{figure*}

\section{Self-Correcting Multi-Agent Prompt Refinement}
\label{sec:method}

In this section, we first present the stage-wise agent-based framework of \name{}. Then, we present the details of each stage.

\subsection{Stage-wise Agent-Based Architecture}

T2V prompts impose heterogeneous reasoning demands that go beyond surface attributes such as length or visual richness. In addition, their difficulty arises in complex scenarios where user intent is under-specified, abstract, or internally entangled. Complex scenarios pose challenges to constructing temporally coherent and semantically faithful video descriptions. A key obstacle is that complex-scenario prompt refinement typically involves multiple coupled constraints. It requires clarifying intent, organizing entities and events into a coherent spatio-temporal structure, and preserving semantic fidelity throughout rewriting. When these requirements are handled implicitly within a single rewrite, the prompt refinement tends to become brittle. As a result, semantic omissions and contradictions may be introduced relative to the user input.

Motivated by this observation, we construct a stage-wise multi-agent pipeline within \name{}, which externalizes the prompt refinement into 5 explicit stages with intermediate representations and verification-driven feedback. The stage-wise multi-agent pipeline incorporates conditional self-correction to achieve excellent T2V generation, in which specialized agents communicate through explicit intermediate representations. As illustrated in Figure~\ref{fig:refinement-pipeline}, \name{} comprises six functional agents organized into two interacting groups. First, the \emph{refinement group} includes a Scenario Router, a Policy Generator, and a Prompt Refiner, which jointly perform scenario-aware strategy selection and policy-conditioned prompt rewriting.
Second, the \emph{verification group} consists of a Semantic Atomizer, an Entailment Validator and a Content Reviser, which collectively enforce semantic fidelity through atom-level verification. In the common case, refinement proceeds forward from routing and policy generation to rewriting and verification.
When semantic fidelity violations are detected, structured feedback from the Entailment Validator selectively triggers revision by Content Reviser, while atomic constraints remain fixed.

\begin{figure*}[t]
    \centering 
    \includegraphics[width=\textwidth]{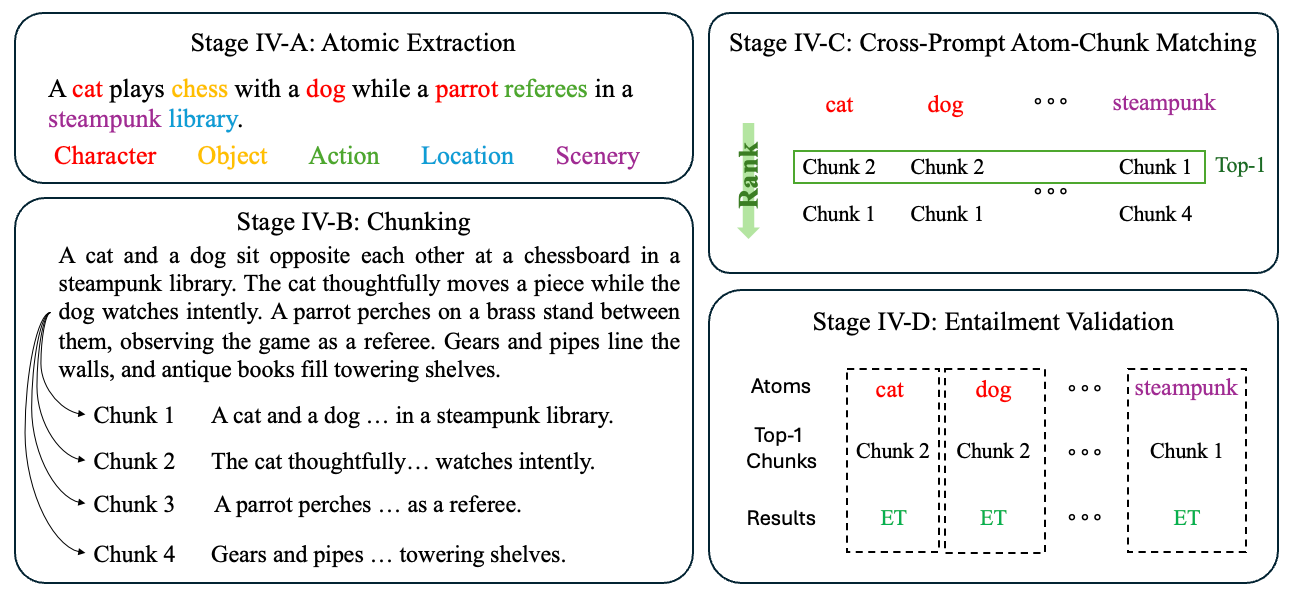}
    \vspace{-7mm}
    \caption[Illustration of Semantic Verification]{
    \textbf{Illustration of the Semantic Verification Stage in \name{}.}
    Given a user input and the corresponding refined prompt, semantic verification is performed in four steps.
    (1) \emph{Atomic Extraction} decomposes the user input into atom elements. 
    (2) \emph{Chunking} segments the refined prompt into semantically coherent evidence units.
    (3) \emph{Atom-Chunk Matching} retrieves the most relevant evidence chunk for each atom. 
    (4) \emph{Entailment Validation} assesses atom-level semantic relations between atoms and evidence chunks.
    Through this design, semantic missing and contradictions in the refined prompt can be detected and subsequently used to trigger downstream revision.
    }
    \vspace{-5mm}
    \label{fig:pipeline_prompt_verification} 
\end{figure*}

\subsection{Stage I: Scenario Tagging for Strategy Routing}
\label{subsec:stage_routing}

\name{} exploits a predefined routing tag set comprising 11 scenario labels of two types, i.e., a non-difficult tag and ten complex-scenario tags: Abstract Descriptions, Complex Spatial Relations, Multi-Element Scenes, Fine-Grained Appearance, Temporal Consistency, Stylistic Hybrids, Causality \& Physics, Camera Motion, Object Interaction, and Scene Transitions (see Appendix~\ref{app-sec:complex_scenario_taxonomy} for formal definitions and representative examples).
The non-difficult tag serves as a conservative fallback for ordinary prompts that do not exhibit a dominant complex constraint, e.g., ``An airplane.'' in VBench~\citep{huang2024vbench},
while the ten complex-scenario tags capture dominant sources of difficulty in complex-scenario T2V prompting and act as routing signals for downstream policy synthesis and prompt refinement.
Formal definitions and representative examples of the ten complex-scenario categories are provided in Appendix~\ref{app-sec:complex_scenario_taxonomy}.

Given user input $P_{\text{user}}$, Scenario Router classifies the input and attributes a corresponding scenario tag $\hat{y}$ based on an LLM (see prompt details in Appendix~\ref{app-subsec:prompt_for_scenario_tagging}). The scenario tag is represented with a one-shot setting selected from the predefined scenario label set.
Then, the routing signal, i.e., Scenario Tag $\hat{y}$, is passed to the Policy Generator for scenario-aware policy synthesis.

\subsection{Stage II: Policy Synthesis for Scenario-Aware Refinement}
\label{subsec:stage_policy} 

\name{} employs the Policy Generator to synthesize a prompt-specific rewriting policy. When receiving user input and scenario tag, i.e., $(P_{\text{user}}, \hat{y})$, from Scenario Router, the Policy Generator outputs a scenario-conditioned rewriting policy $\pi$, which provides explicit guidance on how refinement should be performed under the routed scenario, exploiting an LLM (see prompt details in Appendix~\ref{app:policy_prompts}).

Concretely, the generated policy $\pi$ specifies three aspects.
First, it identifies which implicit constraints should be made explicit (e.g., entities, relations, actions, temporal stages).
Second, it indicates which scenario-relevant modeling competencies should be emphasized (e.g., spatial layout, temporal coherence, physical plausibility, camera motion).
Third, it enforces a conservative fidelity principle that preserves the user-intended meaning and avoids introducing unsupported content.

Scenario Tag $\hat{y}$ is provided as a conditioning context for policy synthesis, while preserving flexibility across prompts.
This design enables refinement strategies to be dynamically synthesized rather than pre-specified.

\subsection{Stage III: Policy-Conditioned Prompt Refinement}
\label{subsec:stage_refine}
Conditioned on the generated policy $\pi$ and the original user input $P_{\text{user}}$, the Prompt Refiner generates a refined prompt $P_{\text{rew}}$ based on an LLM (see prompt details in Appendix \ref{subsec:refine}).
The Refiner rewrites $P_{\text{user}}$ into a clearer and more model-friendly description by elaborating underspecified details, e.g., characters, objects, actions, and settings, while preserving the original intent, tone, and stated content.
Guided by $\pi$, the refiner avoids introducing unsupported elements and produces the refined prompt as a small number of concise sentences, without additional explanations.
By decoupling policy synthesis from prompt rewriting, \name{} encourages faithful execution of explicit constraints and mitigates semantic deviation during refinement.


\begin{table*}[t]
\centering
\caption{Quantitative results (\%) on VBench with LaVie and Wan as T2V backbones. Bold values denote the best performance for each metric under each backbone.}
\vspace{-3mm}
\label{tab:main_results_vbench}
\scalebox{0.88}{
\begin{tabular}{lccccccc}
\toprule
\textbf{Method}
& \shortstack{\textbf{Average}\\\textbf{Score}}
& \shortstack{Aesthetic\\Quality}
& \shortstack{Background\\Consistency}
& \shortstack{Imaging\\Quality}
& \shortstack{Motion\\Smoothness}
& \shortstack{Subject\\Consistency}
& \shortstack{Temporal\\Flickering}
\\
\midrule
LaVie &81.89 &53.52 &95.01 &62.83 &95.15 &90.96 &93.85 \\
LaVie + Open-Sora &81.95&53.67&94.82&62.93&95.28&91.24&93.78\\
LaVie + RAPO &82.80&54.38&95.76&63.06&96.03&93.21&94.35\\
LaVie + \name{} &\textbf{84.56}&\textbf{56.53}&\textbf{97.43}&\textbf{63.87}&\textbf{97.18}&\textbf{95.95}&\textbf{96.39} \\

\midrule
Wan &86.19&63.75&95.41&69.78&97.62&95.12&95.43\\
Wan + Open-Sora &86.27&63.82&95.22&70.05&97.76&95.28&95.52\\
Wan + RAPO &87.43&64.12&97.35&70.99&98.72&96.18&97.23\\
Wan + \name{} &\textbf{88.21}&\textbf{65.19}&\textbf{98.04}&\textbf{71.67}&\textbf{98.98}&\textbf{97.20}&\textbf{98.19}\\

\bottomrule
\end{tabular}
}
\vspace{-2mm}
\end{table*}

\begin{table*}[t]
\centering
\caption{Quantitative comparisons on EvalCrafter. \name{} consistently achieves better results, demonstrating a clear lead on this benchmark.}
\vspace{-3mm}
\label{tab:main_results_evalcrafter}
\scalebox{0.88}{
\begin{tabular}{lccccc}    
\toprule
\textbf{Method} & \textbf{Average} & Motion Quality & Text-Video Alignment & Visual Quality & Temporal Consistency\\
\midrule
LaVie &62.12&53.19&69.60&64.81&60.87\\
LaVie + Open-Sora &62.78&53.07&71.38&65.26&61.41\\
LaVie + RAPO &63.91&53.34&74.38&66.62&61.29\\
LaVie + \name{} 
&\textbf{65.18}&\textbf{53.87}&\textbf{75.42}&\textbf{68.84}&\textbf{62.56}\\
\midrule
Wan               &63.46&53.79 &73.05&65.17&61.85\\
Wan + Open-Sora   &63.95&53.93 &73.84&65.60&62.43\\
Wan +  RAPO       &64.54&{54.26} &{75.13}&{66.08}&{62.71}\\
Wan + \name{} &\textbf{66.74}&\textbf{54.85} &\textbf{76.94}&\textbf{70.23}&\textbf{64.93}\\
\bottomrule
\end{tabular}
}
\vspace{-4mm}
\end{table*}


\subsection{Stage IV: Semantic Verification via Atomization and Entailment Validation}
\label{subsec:stage_verify}

To ensure semantic fidelity, \name{} performs structured verification on the refined prompt.
This stage is realized by Semantic Atomizer and Entailment Validator, which interact through explicit intermediate representations to assess consistency between $P_{\text{user}}$ and $P_{\text{rew}}$. Details are available in Figure~\ref{fig:pipeline_prompt_verification}.

\paragraph{Stage IV-A: Atomic Extraction as Fixed Verification Targets.}
Given the user input $P_{\text{user}}$, Semantic Atomizer extracts a field-wise atom dictionary $\mathcal{D}_\mathcal{A}=\operatorname{Atomizer}(P_{\text{user}})$ based on an LLM (see prompt details in Appendix \ref{subsec:App:Atomization}). This dictionary contains five fields, namely $\texttt{characters}$, $\texttt{objects}$, $\texttt{actions}$, $\texttt{locations}$ and $\texttt{scenery}$. 
For instance, given
\textit{``A cat plays chess with a dog while a parrot referees in a steampunk library''} as user input,
Semantic Atomizer produces 
$\mathcal{D}_{\mathcal{A}}=$\{
\texttt{characters}: [cat, dog, parrot],
\texttt{objects}: [chess],
\texttt{actions}: [plays, referees],
\texttt{locations}: [library],
\texttt{scenery}: [steampunk]
\}. 
We treat $\mathcal{D}_{\mathcal{A}}$ as fixed verification targets throughout subsequent stages. For subsequent matching, we flatten the dictionary into a single list
$\mathcal{A}=\operatorname{Flatten}(\mathcal{D}_{\mathcal{A}})=\{a_1,a_2,\ldots,a_n\}$,
where $a_i$ is an extracted atom element (e.g., $a_1=\text{"cat"}$) and $n$ represents the number of atom elements in $P_{\text{user}}$.

\paragraph{Stage IV-B: Sentence-Level Chunking.}
The refined prompt $P_{\text{rew}}$ is segmented into non-overlapping chunks $\mathcal{C}=\operatorname{Chunk}(P_{\text{rew}})=\{c_1,\dots,c_m\}$ with $m$ referring to the number of chunks in $P_{\text{rew}}$. By default, each sentence forms a chunk. Short sentences are merged with subsequent ones until a length threshold is reached. 

\paragraph{Stage IV-C: Evidence Selection via Atom-Chunk Matching.}
Given atom elements $\{a_i\}_{i=1}^{n}$ and chunks $\{c_j\}_{j=1}^{m}$, we embed atom elements and chunks into a shared semantic space using a lightweight embedding model $f_{\mathcal{E}}(\cdot)$ and compute cosine similarity $s_{ij}=\cos\left(f_{\mathcal{E}}(a_i), f_{\mathcal{E}}(c_j)\right)$.
For atom element $a_i$, corresponding evidence chunk $e_i$ is selected by maximizing the cosine similarity, i.e., $e_i = \arg\max_{j\in\{1,\ldots,m\}} s_{ij}$. 

\paragraph{Stage IV-D: Atom-Level Entailment Validation.}
For each atom element $a_i \in \mathcal{A}$, we pair it with the selected evidence chunk $e_i \in \mathcal{C}$ and invoke Entailment Validator to assess whether the semantics expressed by $a_i$ are supported by $e_i$ with an LLM (see prompt details in Appendix \ref{subsec:App:entailment}).
Formally, the validator generates a ternary decision
$v_i=\operatorname{Validator}(a_i, e_i)$,
where $v_i \in \{\text{ET}, \text{MS}, \text{CT}\}$ denotes the entailment label for atom $a_i$ with respect to chunk $e_i$, i.e., \emph{EnTailment} (ET), \emph{MiSsing} (MS), and \emph{ConTradiction} (CT). 
ET indicates that the semantics of $a_i$ are well supported by $e_i$.
MS refers to the case, in which $e_i$ provides little support on the semantic expression of $a_i$. CT indicates that $e_i$ has opposite semantic meanings compared to $a_i$.

\begin{table*}[t]
\centering
\caption{Quantitative comparisons on T2V-CompBench. \name{} achieves the highest average score.}
\vspace{-3mm}
\label{tab:main_results_t2v_compbench}
\scalebox{0.88}{
\begin{tabular}{lccccc}    
\toprule
\textbf{Method} & \textbf{Average} & Consistent Attribute & Dynamic Attribute & Action Binding & Motion Binding \\
\midrule
LaVie &0.388&0.620&0.232&0.483&0.215\\
LaVie + Open-Sora&0.361&0.532&0.214&0.470&0.226\\
LaVie + RAPO &0.460&0.692&0.267&0.635&0.243\\
LaVie + \name{} &\textbf{0.476}&\textbf{0.704}&\textbf{0.273}&\textbf{0.640}&\textbf{0.285}\\
\midrule
Wan                  &0.454 & 0.694 & 0.263 & 0.591 & 0.269 \\
Wan + Open-Sora      &0.446 & 0.672 & 0.258 & 0.583 & 0.274 \\
Wan + RAPO           &0.495& 0.721 & 0.279 & 0.672 & 0.309 \\
Wan + \name{}        &\textbf{0.523}&\textbf{0.756} & \textbf{0.297} & \textbf{0.691} & \textbf{0.346} \\
\bottomrule
\end{tabular}
}
\vspace{-2mm}
\end{table*}

\begin{table*}[t]
\centering
\caption{
Quantitative results (\%) on T2V-Complexity using VBench metrics with Wan as the T2V backbone.
}
\label{tab:app_t2vcomplexity_vbench_wan}
\vspace{-3mm}
\scalebox{0.9}{
\begin{tabular}{lccccccc}
\toprule
\textbf{Method}
& \shortstack{\textbf{Average}\\\textbf{Score}}
& \shortstack{Aesthetic\\Quality}
& \shortstack{Background\\Consistency}
& \shortstack{Imaging\\Quality}
& \shortstack{Motion\\Smoothness}
& \shortstack{Subject\\Consistency}
& \shortstack{Temporal\\Flickering}
\\
\midrule
Wan         &82.95&56.95&94.83&63.80&96.27&91.28&94.59 \\
Wan + \name{} &85.69&59.32&96.24&67.87&98.87&93.94&97.92 \\
\bottomrule
\end{tabular}
}
\vspace{-3mm}
\end{table*}

\subsection{Stage V: Conditional Revision}
\label{subsec:stage_revision}
To monitor verification quality, we compute the coverage rate $p_\text{ET}$ and contradiction rate $p_\text{CT}$ as diagnostic statistics after atom-level entailment judgments. Specifically, they measure the fractions of atom-chunk pairs labeled as entailment and contradiction, respectively:
\begin{align}
    p_\text{ET}
    &= \frac{1}{|\mathcal{A}|}
    \sum_{a_i\in\mathcal{A}}
    \mathbb{I}\left[\operatorname{Validator}(a_i,e_i)=\text{ET}\right], \\
    p_\text{CT}
    &= \frac{1}{|\mathcal{A}|}
    \sum_{a_i\in\mathcal{A}}
    \mathbb{I}\left[\operatorname{Validator}(a_i,e_i)=\text{CT}\right].
\end{align}
In our framework, $p_\text{ET}$ and $p_\text{CT}$ are used for diagnostic monitoring rather than as soft indicators for revision decisions. We enforce a strict acceptance criterion that the refinement is accepted only if $p_\text{ET}=1$ (100\%) and $p_\text{CT}=0$. Otherwise, Entailment Validator returns structured feedback pinpointing missing or contradicted atoms, which triggers Content Reviser to revise the prompt and produce an updated $P_{\text{rew}}$. The procedure repeats until the acceptance criterion is met or a maximum number of revision rounds is reached. In many cases, the criterion is satisfied immediately, and the framework completes in a single forward pass.


\begin{table}[t]
\centering
\caption{Statistics of complex-scenario prompts across different T2V benchmarks. Here, CS is the abbreviation for complex-scenario.} 
\label{tab:complex_scenario_benchmarks}
\vspace{-3mm}
\resizebox{\linewidth}{!}{
\begin{tabular}{lcccc}
\toprule
T2V Benchmark & Prompts & CS Prompts & CS Proportion & CS Categories \\
\midrule
VBench         & 946  & 311  & 32.9\%  & 9  \\
EvalCrafter    & 700  & 422  & 60.3\%  & 9  \\
T2V-CompBench  & 1400 & 972  & 69.4\%  & 10 \\
T2V-Complexity & 1000 & 1000 & 100.0\% & 10 \\
\bottomrule
\end{tabular}
}
\end{table}

\section{T2V-Complexity Benchmark}

To enable rigorous evaluation of T2V generation under taxonomy-defined complex scenarios, we introduce \emph{T2V-Complexity}. This benchmark contains 1000 user-style prompts, with 100 prompts for each of the ten complex-scenario categories in our taxonomy.
Each prompt corresponds to a target generation task in its designated category while accompanied by expected failure modes, which enables interpretable and fine-grained analysis.

In contrast to prior benchmarks that contain a mixture of ordinary and complex prompts, T2V-Complexity is constructed to focus exclusively on complex-scenario prompts. Table~\ref{tab:complex_scenario_benchmarks} shows that VBench (946 prompts), EvalCrafter (700 prompts) and T2V-CompBench (1400 prompts) contain 32.9\%, 60.3\%, and 69.4\% complex-scenario prompts, respectively. Crucially, T2V-Complexity comprises 1000 prompts that are all complex-scenario prompts and covers 10 complex-scenario categories with a balanced design. This design aligns with our motivation that existing benchmarks may provide limited coverage of certain complex categories and exhibit severe scenario imbalance (also see Appendix~\ref{supp-subsec:analysis_on_scenario_distribution}), which poses challenges for fair and effective evaluation.



\begin{table}
\centering
\caption{Ablation studies of different components in \name{} on VBench.}
\label{tab:ablation_study}
\vspace{-3mm}
\scalebox{0.88}{
\begin{tabular}{lc}
\toprule
\textbf{Method} & \textbf{Average Score} \\
\midrule
\name{}   &88.21\%\\
w/o Scenario Routing & 86.49\% \\
w/o Policy Generation & 87.75\% \\
w/o Verification \& Self-Correction & 87.63\% \\
\bottomrule
\end{tabular}}
\vspace{-5mm} 
\end{table}

\begin{figure*}[t]
    \centering
    \begin{subfigure}[b]{\textwidth}
        \centering
        \includegraphics[width=\textwidth]{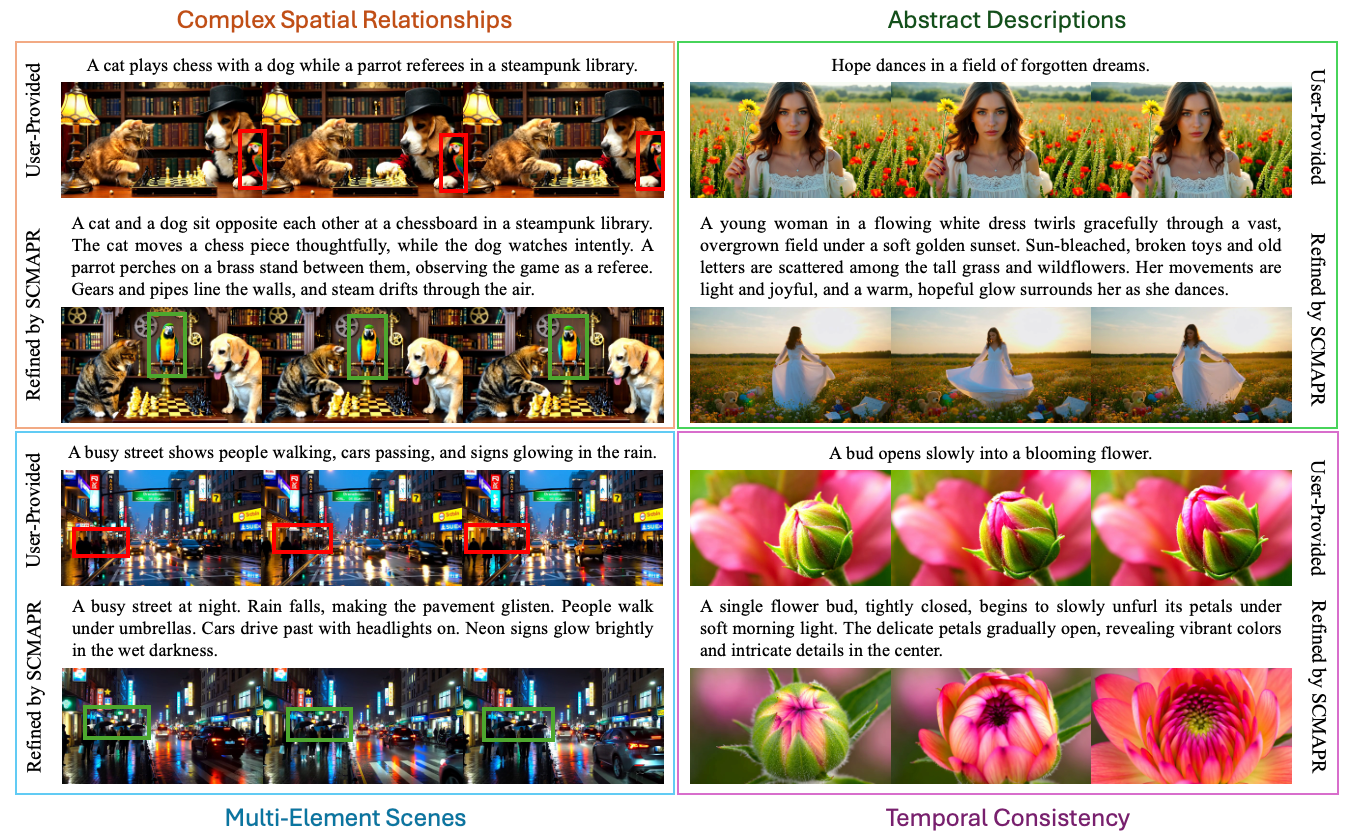}
    \end{subfigure}
    \vspace{-6mm}
    \caption{Comparisons of videos generated using Wan~\citep{wan2025} conditioned on user input and refined prompts from \name{}.}
    \vspace{-4mm}
    \label{fig:scenario_specific_example}
\end{figure*}

\section{Experiments}\label{sec:experiments}

In this section, we present the experimental settings with three benchmarks. Then, we demonstrate the main experimental results. Finally, we show an ablation study.

\subsection{Experimental Setup}

\textbf{Benchmarks: } 
We conduct evaluations on three existing State-Of-The-Art (SOTA) benchmarks, i.e., VBench~\citep{huang2024vbench}, EvalCrafter~\citep{liu2024evalcrafter}, and T2V-CompBench~\citep{sun2025t2v}, to evaluate the quality of T2V generation. In addition, we further evaluate SCMAPR on our T2V-Complexity, which is designed for complex scenarios. All table metrics are averaged over prompts.

\textbf{Baselines:} We compare three SOTA representative prompt refinement approaches in the field of T2V, including the direct prompting (using user input), 
prompt refiner from Open-Sora~\citep{opensora2024}, and RAPO~\citep{gao2025devil}.

\textbf{Implementation:} 
In our experiments, we adopt DeepSeek-V3.2~\citep{deepseek2025deepseekv32} to construct multiple agents. 
In addition, we employ BGE-M3 \cite{chen-etal-2024-m3} as the embedding model in atom-chunk matching.
For video generation, we adopt Wan2.2 (Wan)~\citep{wan2025} and LaVie~\citep{DBLP:journals/ijcv/WangCMZHWYHYYGWSJCLDLQL25} as T2V backbones. 

\subsection{Main Results}\label{subsec:main_results}
As shown in Tables~\ref{tab:main_results_vbench}, \ref{tab:main_results_evalcrafter} and \ref{tab:main_results_t2v_compbench}, we report the quantitative results on three benchmarks, using LaVie and Wan as T2V backbones.
Under both backbones, \name{} consistently achieves the best overall performance, demonstrating its effectiveness for T2V generation. On VBench, \name{} achieves the highest average score of 84.56\% with LaVie and 88.21\% with Wan. Compared with direct prompting, Open-Sora and RAPO, \name{} improves the average score by 2.67\%/2.02\%, 2.61\%/1.94\%, and 1.76\%/0.78\% under LaVie/Wan, respectively.
We can observe similar trends on EvalCrafter, where \name{} achieves the highest average score of 65.18 with LaVie and 66.74 with Wan.
Compared with direct prompting, Open-Sora and RAPO, \name{} improves the average score by 3.06/3.28, 2.40/2.79 and 1.27/2.20 under LaVie/Wan, respectively.
Notably, it demonstrates consistent improvements in T2V alignment and temporal consistency, underscoring its strength in semantic fidelity and dynamic coherence.
When it comes to T2V-CompBench with an emphasis on compositional reasoning, \name{} achieves the best results in consistent attribute binding (up to 0.035 higher), dynamic attribute binding (up to 0.018 higher), action binding (up to 0.019 higher), and motion binding (up to 0.042 higher), demonstrating excellent performance in fine-grained attribute and motion modeling.
In particular, \name{} significantly surpasses RAPO, Open-Sora, and direct prompting by 0.016/0.028, 0.115/0.077, and 0.088/0.069 on T2V-CompBench in terms of average score under LaVie/Wan, respectively.

\textbf{Results on T2V-Complexity.} We further evaluate SCMAPR on T2V-Complexity, which consists exclusively of complex-scenario prompts. As shown in Table~\ref{tab:app_t2vcomplexity_vbench_wan}, adopting the VBench metrics, SCMAPR improves the average score from 82.95\% to 85.69\%, yielding a 2.74\% gain over direct prompting. Moreover, SCMAPR achieves consistent improvements across all six metrics, including aesthetic quality, background consistency, imaging quality, motion smoothness, subject consistency, and temporal flickering. These results further verify that SCMAPR remains effective when evaluation is restricted to exclusively complex-scenario prompts, leading to broadly improved visual quality and temporal stability under challenging scenarios.


To provide an intuitive demonstration on the advantages of \name{}, Figure~\ref{fig:scenario_specific_example} highlights the effectiveness of \name{} across four categories of complex scenarios: complex spatial relations, abstract descriptions, multi-entity scenes and temporal consistency. In the case of complex spatial relations, \name{} achieves two notable improvements over videos generated directly from user input. (1) The parrot is correctly placed at the center (green box) rather than at the edge (red box). (2) The library is rendered with steampunk elements instead of being depicted as a regular library. For abstract descriptions, \name{} produces richer actions and vivid visual dynamics. In multi-entity scenes, \name{} meticulously introduces umbrellas for pedestrians in rainy weather (green box). Finally, under temporal consistency, while direct prompting fails to present a fully blossomed flower, \name{} generates the complete blooming process, resulting in a fully blossomed flower video. 
Beyond quantitative gains, \name{} demonstrates effectiveness in reducing hallucinations, especially under abstract or unspecified user input.
\name{} turns vague intentions into grounded visual constraints and coherent descriptions, which can reduce spurious content (e.g., unintended faces or irrelevant objects) and improve concept faithfulness (see details in Appendix~\ref{supp-sec:hallucination_elimination}).


\subsection{Ablation Study}
We conduct ablation experiments on key components of the agentic refinement framework on VBench.
Table~\ref{tab:ablation_study} shows that the full \name{} achieves the best performance.
Removing scenario routing leads to the largest degradation (1.72\%), indicating the necessity of category-specific rewriting policies.
Disabling policy generation yields a smaller drop (0.46\%), suggesting that prompt-specific policy synthesis provides additional gains beyond routing alone.
Finally, removing the verification and self-correction also noticeably degrades performance (0.58\%), verifying that atom-level verification and conditional correction contribute to maintaining semantic fidelity.

\section{Conclusion}
In this paper, we propose \emph{Self-Correcting Multi-Agent Prompt Refinement} (\name{}), i.e., a structured framework for refining text input under complex T2V scenarios.
\name{} formulates prompt refinement as a stage-wise multi-agent framework, in which six specialized agents collaboratively operate across five functional stages, including scenario routing, policy synthesis, prompt refinement, semantic verification and conditional revision. By combining taxonomy-grounded routing with atom-level entailment-based verification, \name{} enables targeted prompt refinement while preserving complete user intent through conditional self-correction. 
Extensive experiments on VBench, EvalCrafter and T2V-CompBench demonstrate that \name{} consistently improves text-video alignment and overall generation quality under complex scenarios, achieving up to 2.67\% and 3.28 gains in average score on VBench and EvalCrafter, respectively, and up to 0.028 improvement on T2V-CompBench compared with state-of-the-art baseline approaches.

\section*{Limitations}
\name{} is a training-free and model-agnostic framework that improves text-to-video generation under complex scenarios via stage-wise multi-agent collaboration.
Nevertheless, several limitations remain. First, the multi-agent design introduces additional inference overhead, especially in the semantic verification stage involving atomic extraction and entailment judgment. Second, the effectiveness of scenario routing and semantic verification depends on the reasoning capability of the instruction-tuned LLM, which may lack adapted fine-tuning \cite{liu2024fisher} or reinforcement learning \cite{zhu2025carft,zhu2025sgdpo}. Third, \name{} adopts a predefined set of scenario tags as lightweight routing signals, which capture major orthogonal sources of difficulty in T2V generation but do not preclude the incorporation of additional categories as new challenges arise. Finally, in this work, we only focus on text inputs, and extending the framework to multimodal inputs requires further adaptation \cite{liu2025nexus}.

\section*{Ethical Considerations}
This work aims to improve prompt refinement for text-to-video generation. As a training-free framework, \name{} does not introduce new generative capabilities beyond the T2V model and LLM. Nevertheless, like other prompt-based systems, it may pose potential risks of misuse if deployed without appropriate safeguards. T2V-Complexity has been manually inspected to ensure that each benchmark item does not contain any information that names or uniquely identifies individual people. In practice, responsible deployment should rely on the safety policies and content filtering mechanisms of the base models, together with standard safeguards such as prompt moderation and output filtering.

\section*{Acknowledgements}
Part of the work is supported by National Natural Science Foundation
of China (No. 72293583), Major Project for Tackling Key Problems in Philosophy and Social Science Research of the Ministry of Education (grant number: 24JZD040), and Consulting Project of the Chinese Academy of Engineering (No. 2025-XZ-08).

\bibliography{custom}

\appendix
\setcounter{figure}{0}
\renewcommand*{\thefigure}{A.\arabic{figure}}

\setcounter{table}{0}
\renewcommand*{\thetable}{A.\arabic{table}}

\setcounter{equation}{0}
\renewcommand\theequation{A.\arabic{equation}}

\clearpage

\section{Complex Scenario Taxonomy}
\label{app-sec:complex_scenario_taxonomy}
In this section, we present the ten-category complex-scenario taxonomy utilized in our benchmark. 
Section~\ref{supp-subsec:overview_design} introduces the motivation of the taxonomy, and summarizes the design principles and organizing dimensions.
Section~\ref{supp-subsec:defs} provides formal definitions and representative examples for all ten complex scenarios.
Section~\ref{supp-subsec:discussion_ext} discusses the extensibility of the taxonomy.
Section~\ref{supp-subsec:primary_label} specifies the primary-label annotation rule and the category-level non-overlap rationale.

\subsection{Overview and Design Principles}\label{supp-subsec:overview_design}
To systematically characterize user inputs that are particularly challenging for current Text-to-Video (T2V) systems, we introduce a ten-category taxonomy of \emph{complex scenarios}, exploiting the same category definitions as those in Section \ref{sec:intro}. Rather than a purely heuristic list, the taxonomy is grounded in:
(i) recurring failure patterns reported in recent T2V benchmarks and systems, (ii) established notions from video understanding and vision-language reasoning (e.g., semantic grounding, spatial layout, temporal coherence, and physical plausibility), and (iii) principles from cinematography and visual storytelling, e.g., camera motion, scene transitions, and stylistic control.

The taxonomy is designed to be approximately mutually exclusive at the category level, while jointly covering a broad spectrum of practically important difficulties in T2V generation.
It is also explicitly extensible, allowing new categories to be covered as additional patterns of systematic misalignment emerge.

From a conceptual perspective, the ten categories can be viewed as spanning several orthogonal sources of difficulty:
\emph{semantic abstraction} (e.g., Abstract Descriptions),
\emph{spatial and compositional structure} (e.g., Complex Spatial Relations, Multi-Element Scenes),
\emph{appearance fidelity} (e.g., Fine-Grained Appearance),
\emph{temporal and physical dynamics} (e.g., Temporal Consistency, Causality \& Physics, Object Interaction),
and \emph{cinematic control} (e.g., Camera Motion; Scene Transitions and Stylistic Hybrids).
Each prompt in our benchmark is assigned exactly one primary category according to its dominant source of difficulty for T2V models. Table~\ref{supp-tab:taxonomy_summary} summarizes the ten categories and their core difficulties. 

\begin{table*}[t]
\centering
\small
\begin{tabular}{p{3.3cm} p{6.2cm} p{5.4cm}}
\toprule
\textbf{Taxonomy Category} & \multicolumn{1}{c}{\textbf{Core Difficulty}} & \multicolumn{1}{c}{\textbf{Non-overlap Rationale}} \\
\midrule
Abstract Descriptions
& Mapping non-visual, metaphorical, or symbolic language to coherent visual realizations.
& Not tied to concrete visual structures; focuses on metaphorical or symbolic interpretation. \\
Complex Spatial Relations
& Satisfying explicit geometric constraints (e.g., left, right, between or center) with stable layout, depth perception and occlusion reasoning.
& Independent of entity count, interaction, or physics; concerns layout, depth, and occlusion. \\
Multi-Element Scenes
& Preserving all salient elements, entity counts, and scene completeness under high visual density without omissions or unintended merging.
& Requires stable configurations among multiple entities; does not involve contact or force dynamics. \\
Fine-Grained Appearance
& Sustaining high-frequency details such as textures, text, and identity-specific features across frames.
& Addresses textures, identity, and small-scale details; orthogonal to spatial, temporal, and physical reasoning. \\
Temporal Consistency
& Avoiding temporal drift, flickering, or motion discontinuities in cross-frame evolution.
& Focuses on continuity of motion and appearance across frames; unrelated to layout or object count. \\
Stylistic Hybrids
& Enforcing coherent visual style under mixed or evolving artistic domains without degradation.
& Fully decoupled from spatial, physical, or interaction constraints; concerns aesthetic specification. \\
Causality \& Physics
& Producing physically plausible motions and cause--effect dynamics consistent with real-world mechanics.
& Evaluates plausible dynamics and cause--effect structure; distinct from contact-based interaction. \\
Camera Motion
& Generating stable and continuous viewpoint trajectories (e.g., pan, tilt, zoom, orbit) without spatial distortion or motion artifacts.
& Governs viewpoint trajectories such as pans, zooms, and orbits; does not affect in-scene relations. \\
Object Interaction
& Modeling contact-driven dynamics (touch, grasp, collision, pouring) with force-dependent responses and interaction-induced occlusion changes over time.
& Involves contact, manipulation, and inter-entity dependencies; independent of static spatial layout. \\
Scene Transitions
& Managing multi-shot coherence with valid cuts, transitions, and scene-level structural continuity.
& Pertains to shot-level editing and transitions; unrelated to within-shot visual modeling. \\
\bottomrule
\end{tabular}
\caption{Summary of the ten complex-scenario taxonomy categories, including each category's core difficulty for text-to-video generation and the rationale used to ensure non-overlapping primary-label annotation.}
\label{supp-tab:taxonomy_summary}
\end{table*}

\subsection{Definitions of 10 Complex Scenarios}\label{supp-subsec:defs}
\paragraph{Category 1: Abstract Descriptions.}
User inputs in this category describe abstract, metaphorical, or non-physical concepts that lack a direct visual referent, such as emotions, mental states, or symbolic ideas (e.g., \emph{hope}, \emph{nostalgia}, \emph{loneliness}).  
Formally, a user input belongs to this category if its core semantic content cannot be inferred through literal object depiction alone and instead requires symbolic instantiation, personification, or scene-level metaphor.

Example:  
\emph{Hope dances in a field of forgotten dreams.}

Unlike concrete object prompts, these descriptions require T2V models to determine how to visually instantiate abstract intent through color palettes, motion patterns, atmosphere, or narrative cues. As a result, semantic grounding becomes the dominant challenge.

\paragraph{Category 2: Complex Spatial Relations.}
This category includes user inputs that explicitly specify relative spatial arrangements among entities, typically via prepositions or geometric constraints (e.g., \emph{behind}, \emph{between}, \emph{above}, \emph{surrounding}).  
A user input is assigned to this category when correctness critically depends on satisfying these spatial relations, regardless of the number of entities involved.

Example:  
\emph{A cat plays chess with a dog while a parrot hovers above them in the center of a steampunk library.}

Typical failure modes include incorrect placement, depth inversion, occlusion errors, or spatial drift across frames, reflecting limitations in 3D layout reasoning and viewpoint consistency.

\paragraph{Category 3: Multi-Element Scenes.}
User inputs in this category describe scenes containing multiple objects or entities whose \emph{counts} and \emph{overall layout} must remain stable throughout the video.  
Formally, a user input belongs to this category when it involves multiple distinct entities ($|E| \geq 3$) and requires preserving their presence, number, and coarse spatial configuration without collapse or omission.

Example: 
\emph{Ten people at a festival, each wearing a different costume, under fireworks.}

This category subsumes explicit numerical constraints (e.g., \emph{exactly five objects}), as errors in counting and entity disappearance are common failure modes in multi-object layouts under temporal generation.

\paragraph{Category 4: Fine-Grained Appearance.}
This category captures user inputs where correctness hinges on subtle local visual details, such as textures, small text, facial expressions, material properties, or identity-specific features.  
A user input is assigned to this category when these fine-grained attributes are semantically essential and sensitive to minor deviations.

Example:  
\emph{A close-up of a book cover clearly showing the title \textit{Deep Learning 101} in bold letters.}

In T2V generation tasks, maintaining such details consistently across frames is challenging due to resolution limits, temporal noise, and identity drift.

\paragraph{Category 5: Temporal Consistency.}
User inputs in this category require coherent temporal evolution across frames, including smooth motion, stable appearance, and consistent state progression over time.  
A user input is categorized as Temporal Consistency when violations such as flickering, discontinuous motion, or appearance drift undermine semantic correctness.

Example:   
\emph{A flower bud slowly opens into full bloom at sunrise.}

Unlike static images, T2V models must ensure continuity across frames, making temporal alignment and long-range consistency the dominant difficulty.

\paragraph{Category 6: Stylistic Hybrids.}
This category includes user inputs that require blending multiple heterogeneous visual styles or artistic domains within a single coherent video.  
A user input is assigned to this category when two or more distinct style descriptors (e.g., \emph{oil painting} and \emph{cyberpunk}) must coexist without collapsing into a single dominant style.

Example:  
\emph{A medieval castle illuminated by neon cyberpunk signs in the style of Van Gogh.}

The challenge lies in enforcing style consistency over time while preserving scene structure and motion realism.

\paragraph{Category 7: Causality \& Physics.}
User inputs in this category describe events governed by physical laws or explicit cause-effect relationships, such that correctness cannot be judged from isolated frames alone.  
Formally, a user input belongs to this category when it specifies a causal chain (e.g., event $A$ causes event $B$) or requires physically plausible dynamics.

Example:  
\emph{A glass is knocked off the table, falls to the floor, and shatters into pieces.}

Common failure modes include missing effects, reversed causality, or physically implausible trajectories.

\paragraph{Category 8: Camera Motion.}
This category covers prompts where the primary difficulty lies in executing continuous camera movements, such as pans, tilts, zooms, or orbits.  
A user input is categorized as Camera Motion when camera trajectory, rather than object motion, is the main constraint.

Example:
\emph{A slow pan from left to right across a crowded marketplace.}

T2V models generally ignore or only partially realize such directives, leading to unstable or unintended viewpoints.

\paragraph{Category 9: Object Interaction.}
User inputs in this category involve explicit physical or functional interactions among entities, such as contact, manipulation, force application, or occlusion changes.  
A user input belongs to this category when modeling inter-object dynamics is essential beyond static layout or mere co-presence.

Example:  
\emph{A person picks up a cup, pours water into it, and places it back on the table.}

These scenarios require precise temporal coordination and interaction-aware dynamics, which remain challenging for diffusion-based video generation models.

\paragraph{Category 10: Scene Transitions.}
This category captures prompts that require coherent multi-shot structure, including cuts, transitions, or scene-level progression across distinct shots. 
A prompt is assigned here when semantic correctness depends on valid transitions rather than within-shot continuity.

Example:   
\emph{The scene cuts from a busy city street to a quiet room at night.}

Failures often arise as abrupt visual discontinuities, invalid transitions, or loss of narrative coherence across shots.

\subsection{Discussion and Extensibility}\label{supp-subsec:discussion_ext}
Although the ten categories above cover a broad range of complex scenarios observed in practice, the taxonomy is explicitly designed to be extensible.
New categories can be added along the same organizing principles (semantic abstraction, compositional structure, temporal-causal dynamics, stylistic and cinematic control) as future T2V research reveals additional, systematically recurring difficulty types, such as long-horizon narrative coherence or character identity consistency across shots.
In our benchmark, each prompt is annotated with exactly one primary category to enable per-scenario analysis, while secondary tags can be attached when multiple difficulties co-occur.
This provides a structured foundation for evaluating and comparing T2V systems under controlled, interpretable dimensions of prompt complexity.

\subsection{Primary-label Annotation Rule}\label{supp-subsec:primary_label}
Each benchmark prompt is assigned exactly one \emph{primary} category based on its dominant source of difficulty.
If multiple difficulties co-occur, secondary tags are recorded but excluded from the main evaluation. 

To ensure that primary labels are mutually exclusive at the category level, we summarize the non-overlap rationale for each category in Table~\ref{supp-tab:taxonomy_summary}.

\begin{table*}[t]
\centering
\caption{Time cost distribution of each operation under different numbers of GPUs for T2V generation.}
\label{tab:time_cost_distribution}
\begin{tabular}{lccccc}
\toprule
Operation & Duration (s) & 1 GPU (\%) & 2 GPUs (\%) & 4 GPUs (\%) & 8 GPUs (\%) \\
\midrule
Scenario Routing        & 2.2  & 0.44\% & 0.85\% & 1.42\% & 2.11\% \\
Policy Generation       & 5.9  & 1.19\% & 2.28\% & 3.82\% & 5.65\% \\
Prompt Refinement       & 2.8  & 0.57\% & 1.08\% & 1.81\% & 2.68\% \\
Entailment Verification & 14.7 & 2.97\% & 5.68\% & 9.51\% & 14.07\% \\
Content Revision        & 1.9  & 0.38\% & 0.74\% & 1.23\% & 1.82\% \\
Video Generation        & \textemdash & 94.44\% & 89.36\% & 82.20\% & 73.68\% \\
\bottomrule
\end{tabular}
\end{table*}

\begin{table}[t]
\centering
\caption{Time cost of video generation under different numbers of GPUs.}
\label{tab:video_generation_duration}
\begin{tabular}{lc}
\toprule
Operation & Duration (s) \\
\midrule
Video Generation (1 GPU) & 467 \\
Video Generation (2 GPUs) & 231 \\
Video Generation (4 GPUs) & 127 \\
Video Generation (8 GPUs) & 77 \\
\bottomrule
\end{tabular}
\end{table}

\section{Time Consumption}
We profile SCMAPR on the same GPU server setup used in our experiments and report detailed stage-wise time consumption below. The results are demonstrated in Table~\ref{tab:time_cost_distribution} and Table~\ref{tab:video_generation_duration}. 

\textbf{(1) The text-side overhead constitutes only a small fraction of the end-to-end runtime.}
Text-side time consumption is small relative to video generation. 
We perform SCMAPR on the same GPU server used in our experiments. 
On average, the text-side optimization in SCMAPR costs 27.5 seconds per user input, including Scenario Routing (2.2 s), Policy Generation (5.9 s), Prompt Refinement (2.8 s), Entailment Verification (14.7 s), and Content Revision (1.9 s). Entailment Verification is the largest contributor (14.7 s), as it performs structured atom-level checking.

The text-side optimization cost (27.5 s) is independent of GPU parallelism, while video generation scales strongly with the number of GPUs. Video generation takes 467 seconds on a single GPU, and decreases to 231, 127, and 77 seconds when using 2, 4, and 8 GPUs, respectively. Accordingly, the entire 5-stage text pipeline constitutes only 5.56\%, 10.64\%, 17.80\% and 26.32\% of the end-to-end runtime under 1, 2, 4 and 8 GPUs.
When considering the full end-to-end pipeline, SCMAPR accounts for only 5.56\% of total runtime under a single GPU and 10.64\% under two GPUs. Although the relative proportion increases to 17.80\% and 26.32\% with 4 and 8 GPUs, this change is solely due to the accelerated video generation stage rather than any additional overhead from SCMAPR itself. In practical single-video deployment scenarios, where 1–2 GPUs are most common, the added cost therefore remains within 5–11\% of total runtime, and the dominant computational burden consistently lies in the T2V generation backbone rather than the multi-agent reasoning process. Overall, although SCMAPR involves multiple LLM calls, its text-side time consumption remains relatively small compared to the time spent on T2V generation, which is the dominant component in practice.

\textbf{(2) Reducing regeneration cycles can enhance practical efficiency.}
From a practical usage perspective, SCMAPR can reduce overall user time despite introducing a modest upfront optimization cost. When structured prompt refinement is not available, users may frequently encounter misalignment or logical inconsistencies in the generated video, which leads them to manually adjust the prompt and regenerate the video. Since each regeneration typically requires 77-467 seconds depending on GPU configuration, even a single additional iteration is substantially more expensive than the average 27.5 seconds required by SCMAPR for text-side optimization before generation. By improving alignment and consistency in advance, SCMAPR reduces the likelihood of repeated regeneration cycles. As a result, the modest text-side overhead can translate into net time savings in realistic workflows.

\begin{figure*}[t]
    \centering
    \begin{subfigure}[b]{\textwidth}
        \centering
        \includegraphics[width=\textwidth]{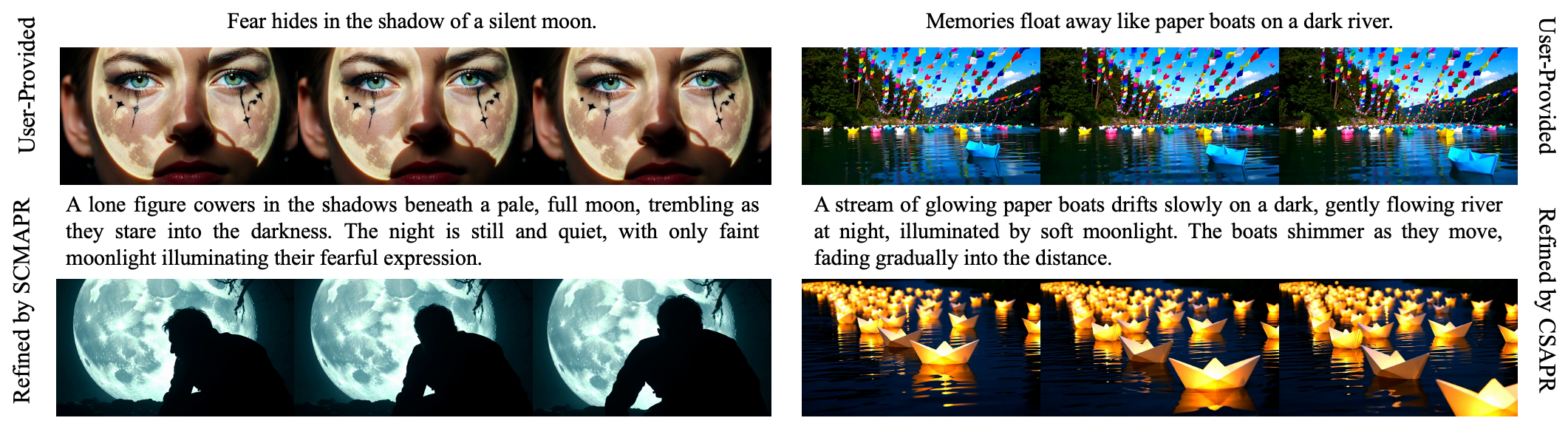}
    \end{subfigure}
    \vspace{-4mm}
    \caption{Examples of hallucination elimination after prompt refinement by \name{}.}
    \label{fig:example_hallucination_additional}
\end{figure*}

\begin{figure*}[!ht]
    \centering
    \begin{subfigure}[b]{\textwidth}
        \centering
        \includegraphics[width=\textwidth]{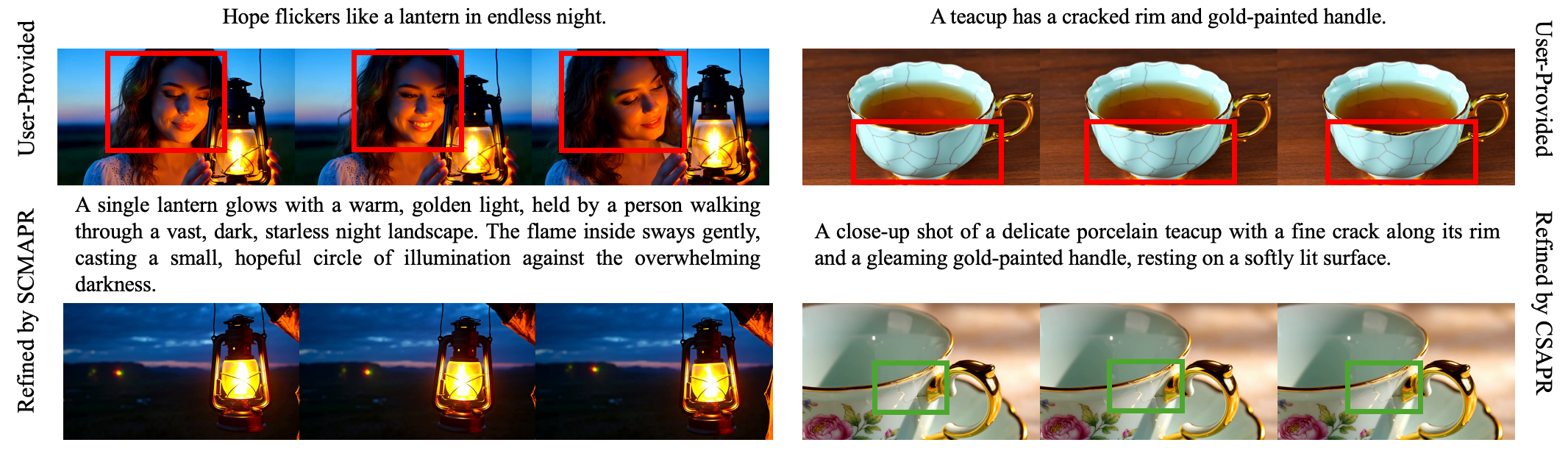}
    \end{subfigure}
    \caption{Examples of hallucination elimination after prompt refinement by \name{}. The hallucinated information is highlighted in red, and its elimination or correction is marked in green.}
    \label{fig:example_hallucination}
\end{figure*}

\section{Hallucination Elimination}
\label{supp-sec:hallucination_elimination}

Figure~\ref{fig:example_hallucination_additional} presents examples where simple prompt descriptions lead to hallucinations with the original user input while \name{} can reduce hallucinations. In the left panel, the original abstract input causes the T2V model to misinterpret the user intent, resulting in an image of a ghostly face illuminated by moonlight. In contrast, the prompt refined by \name{} provides a concrete depiction aligned with the abstract prompt. Therefore, the T2V model can understand the desired content better and then eliminate hallucinations with the refined prompts of \name{}. In the right panel, the output based on the original prompt fails to reflect the concept of a dark river, instead producing dense arrays of small colorful flags irrelevant to the prompt. In comparison, the prompt refined by \name{} conveys a coherent visual narrative that captures the fading of memories and the lingering attachment to the past.

Figure~\ref{fig:example_hallucination} presents examples where simple prompt descriptions lead to hallucinations. In the left panel, although the original input does not specify the presence of a human face, the generated video still contains a woman's face due to spurious correlations in the training data, where scenes with lamps often occur simultaneously with humans. In contrast, the \name{}-refined prompt emphasizes scenery, visual context, and atmosphere, resulting in the generated video that is more aligned with the user intent. In the right panel, the user input requests a bowl with a cracked rim, while the T2V model produces a bowl with grid-like decorations and even unmentioned tea. In comparison, the video generated from the \name{}-refined prompt yields a bowl with the desired cracks (highlighted in green).

\begin{figure*}[!h]
  \centering
  \captionsetup[subfigure]{justification=centering} 
  \begin{subfigure}[b]{0.58\textwidth}
    \centering
    \includegraphics[width=\linewidth]{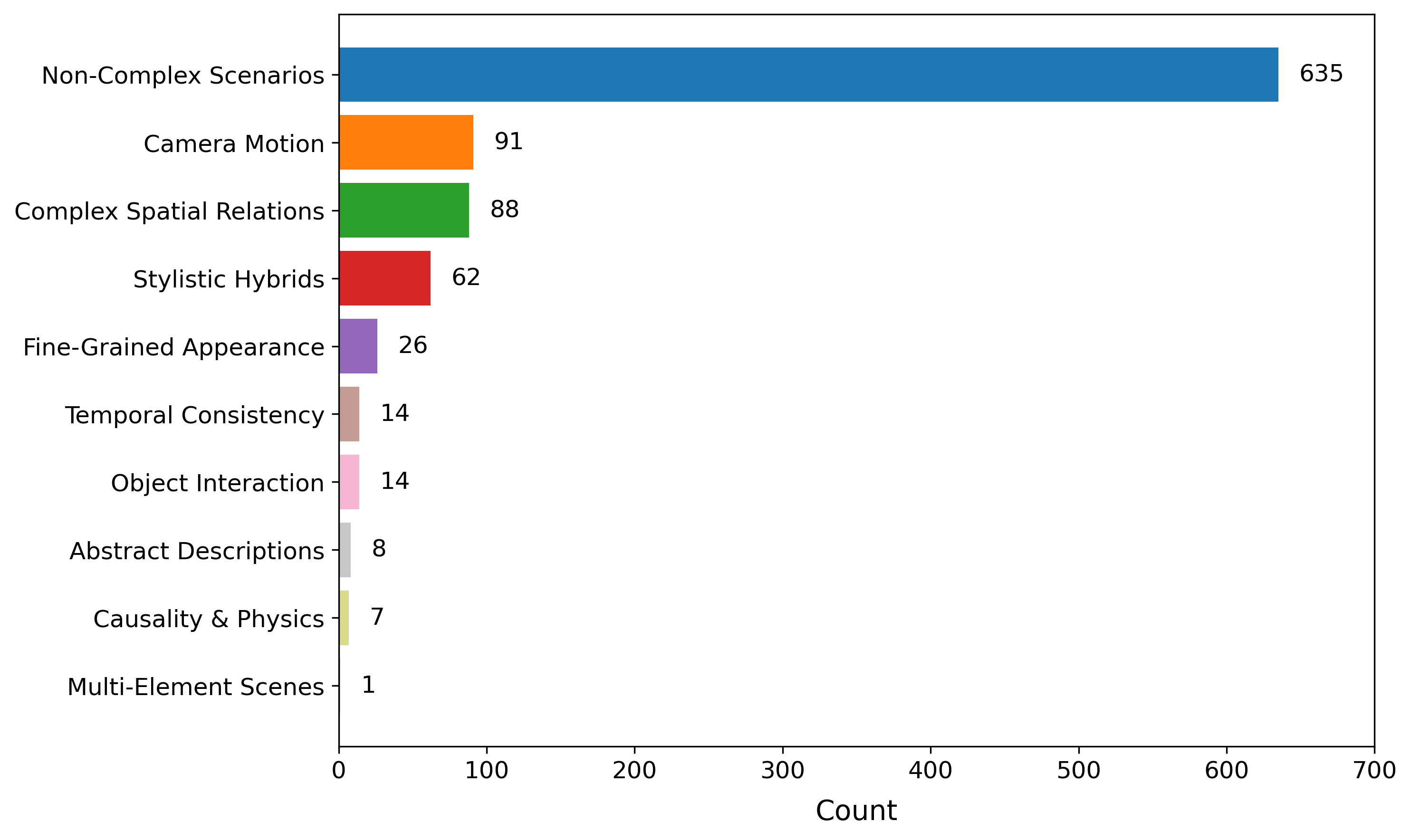}
    \caption{Category distribution of complex scenarios.}
  \end{subfigure}
  \begin{subfigure}[b]{0.38\textwidth}
    \centering
    \includegraphics[width=\linewidth]{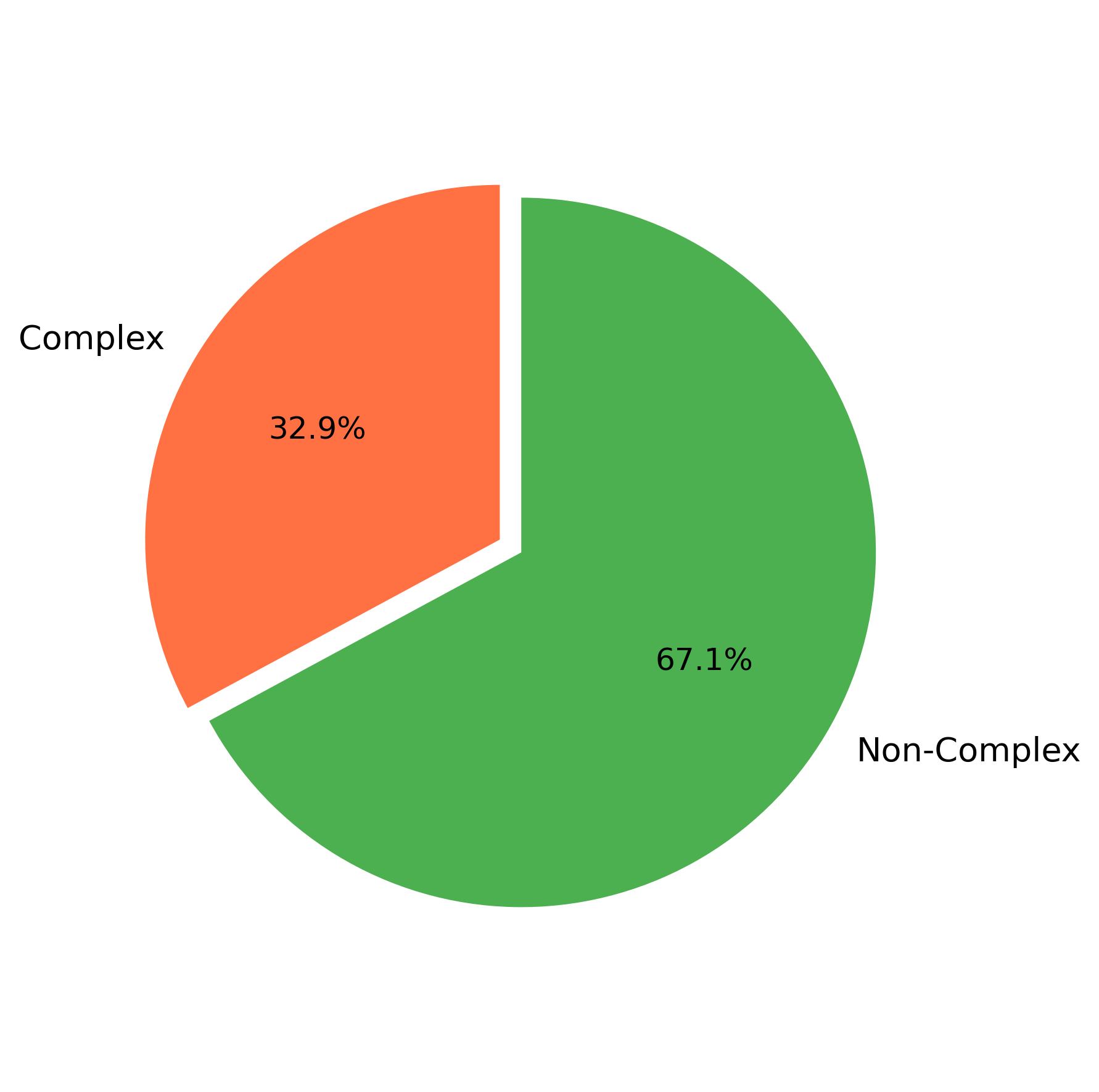}
    \caption{Proportion of complex vs. non-complex.}
  \end{subfigure}
  \caption{Distribution of complex scenarios across VBench.}
  \label{supp-fig:distribution_vbench}
\end{figure*}

\begin{figure*}[!h]
  \centering
  \captionsetup[subfigure]{justification=centering} 
  \begin{subfigure}[b]{0.58\textwidth}
    \centering
    \includegraphics[width=\linewidth]{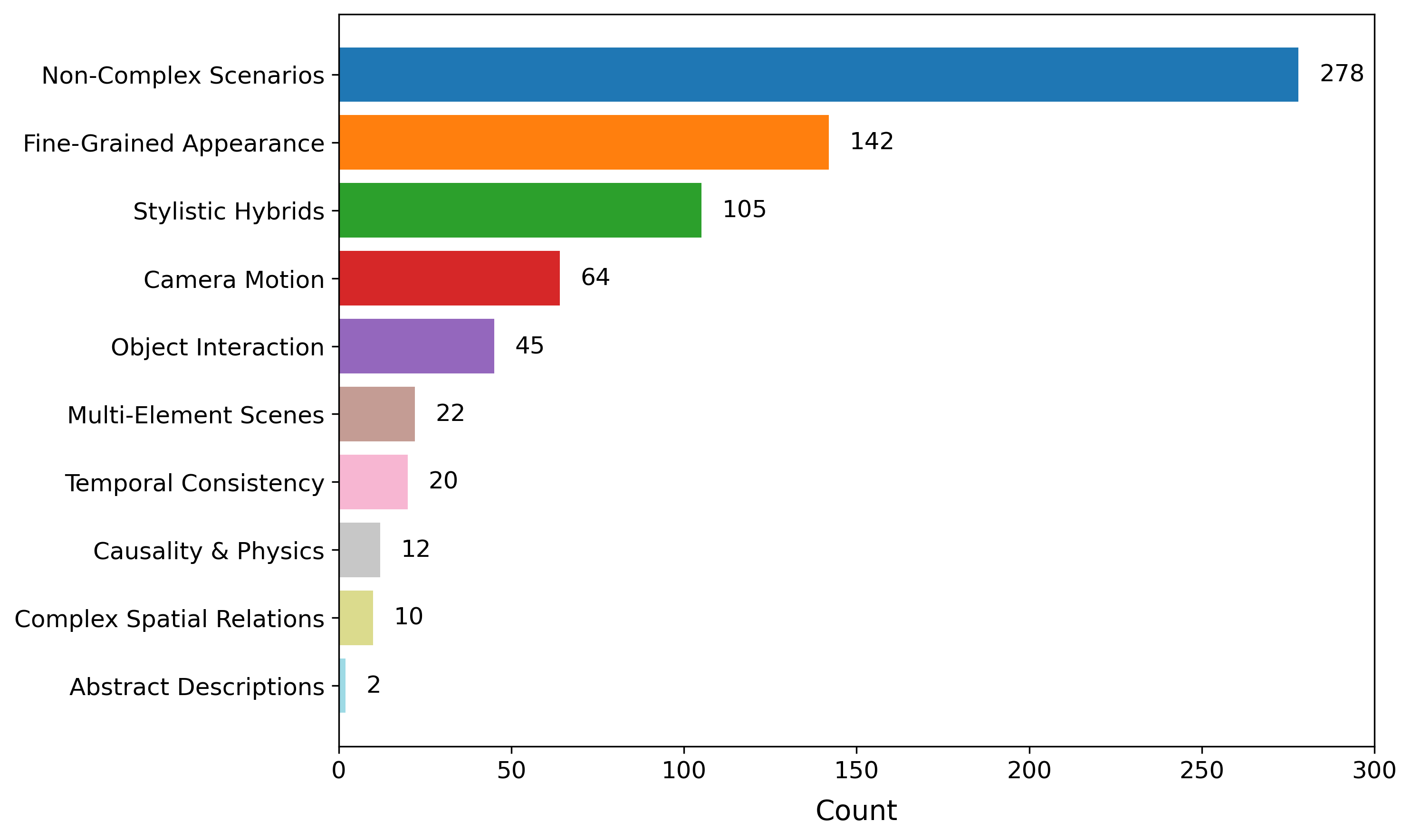}
    \caption{Category distribution of complex scenarios.}
  \end{subfigure}
  \begin{subfigure}[b]{0.38\textwidth}
    \centering
    \includegraphics[width=\linewidth]{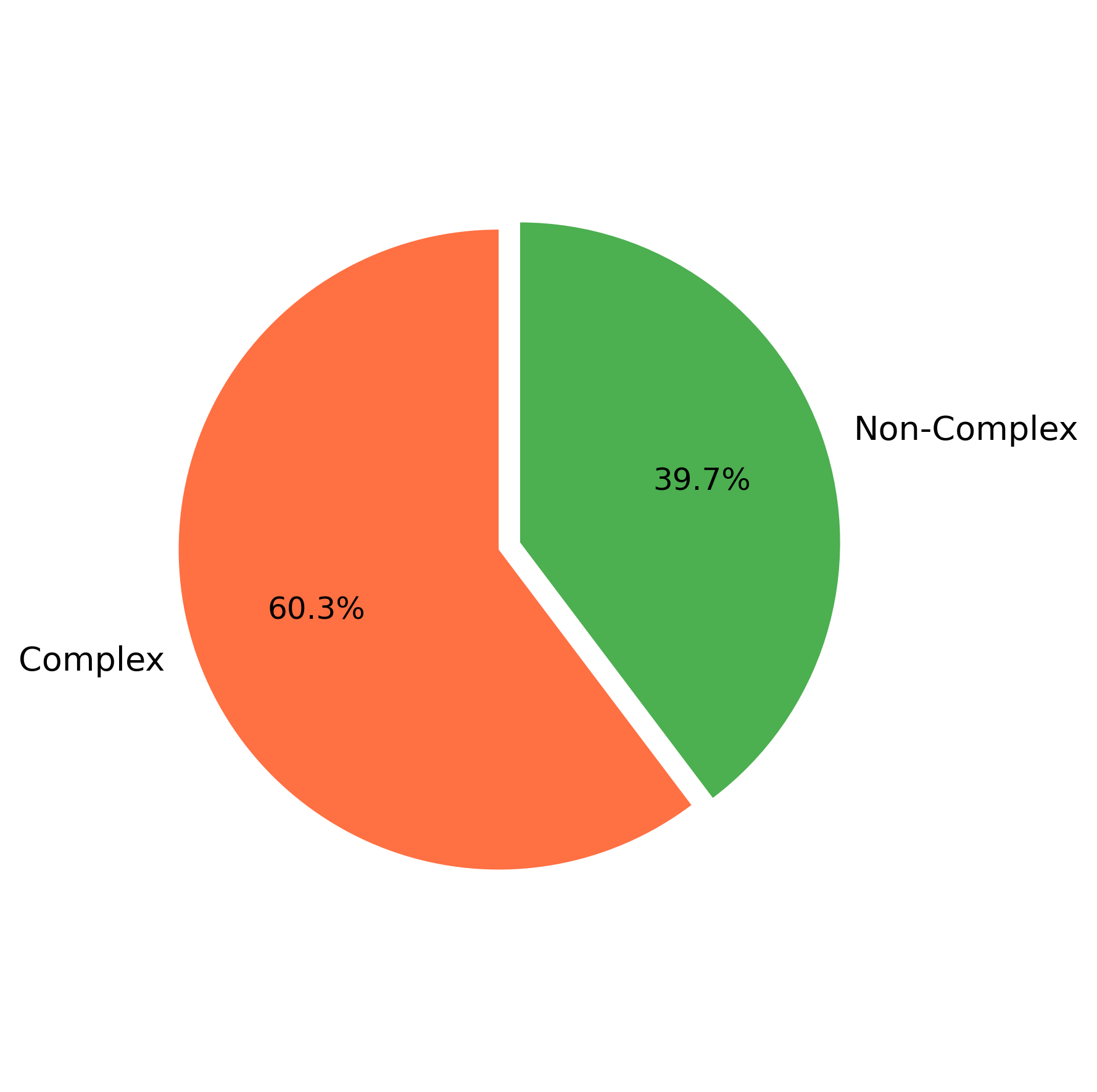}
    \caption{Proportion of complex vs. non-complex.}
  \end{subfigure}
  \caption{Distribution of complex scenarios across EvalCrafter.}
  \label{supp-fig:distribution_evalcrafter}
\end{figure*}

\begin{figure*}[!h]
  \centering
  \captionsetup[subfigure]{justification=centering} 
  \begin{subfigure}[b]{0.58\textwidth}
    \centering
    \includegraphics[width=\linewidth]{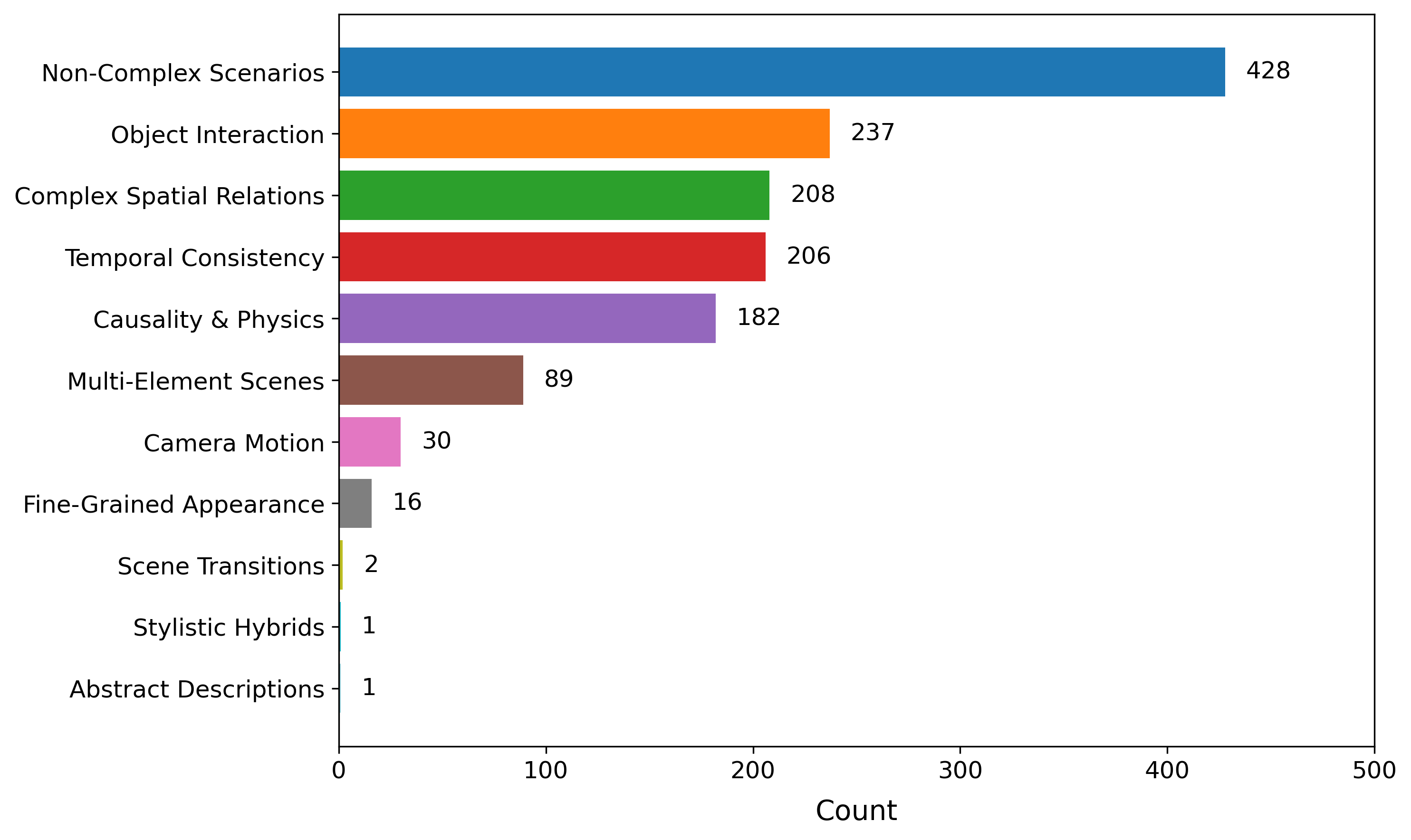}
    \caption{Category distribution of complex scenarios.}
  \end{subfigure}
  \begin{subfigure}[b]{0.38\textwidth}
    \centering
    \includegraphics[width=\linewidth]{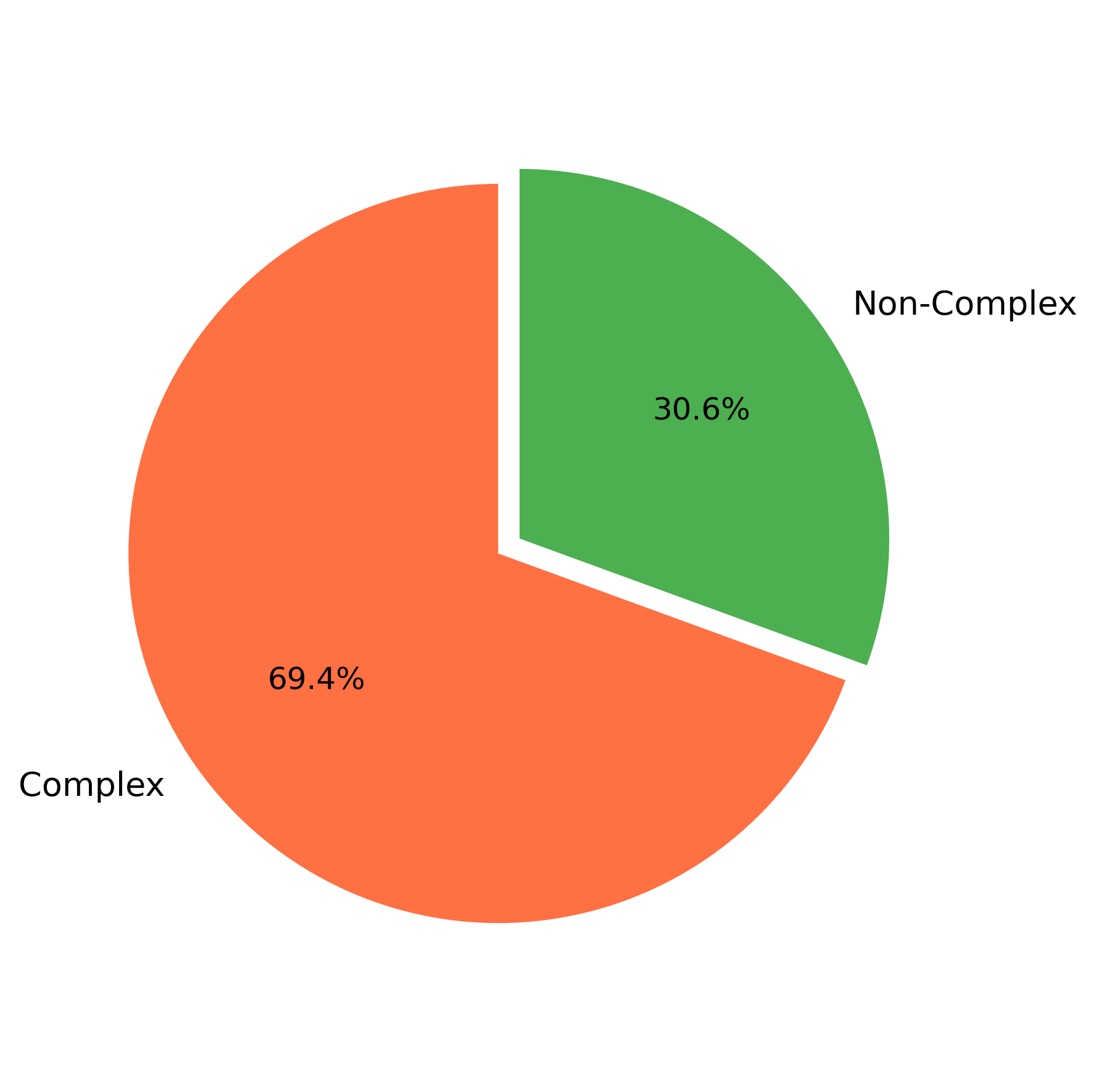}
    \caption{Proportion of complex vs. non-complex.}
  \end{subfigure}
  \caption{Distribution of complex scenarios across CompBench.}
  \label{supp-fig:distribution_compbench}
\end{figure*}

\begin{figure*}[!h]
  \centering
  \captionsetup[subfigure]{justification=centering}
  \begin{subfigure}[b]{0.7\textwidth}
    \centering
    \includegraphics[width=\linewidth]{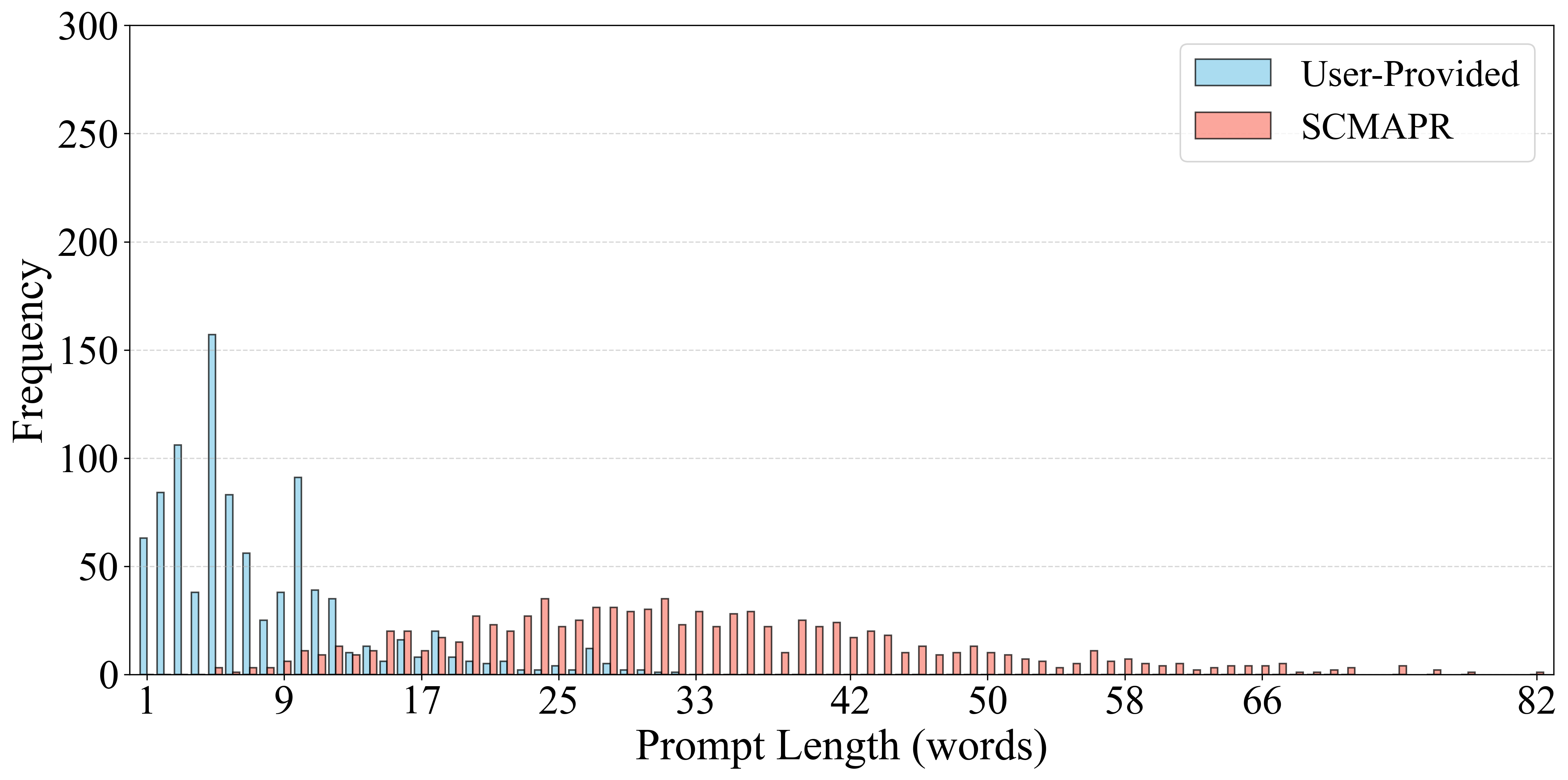}
    \caption{VBench}
  \end{subfigure}

\begin{subfigure}[b]{0.7\textwidth}
    \centering
    \includegraphics[width=\linewidth]{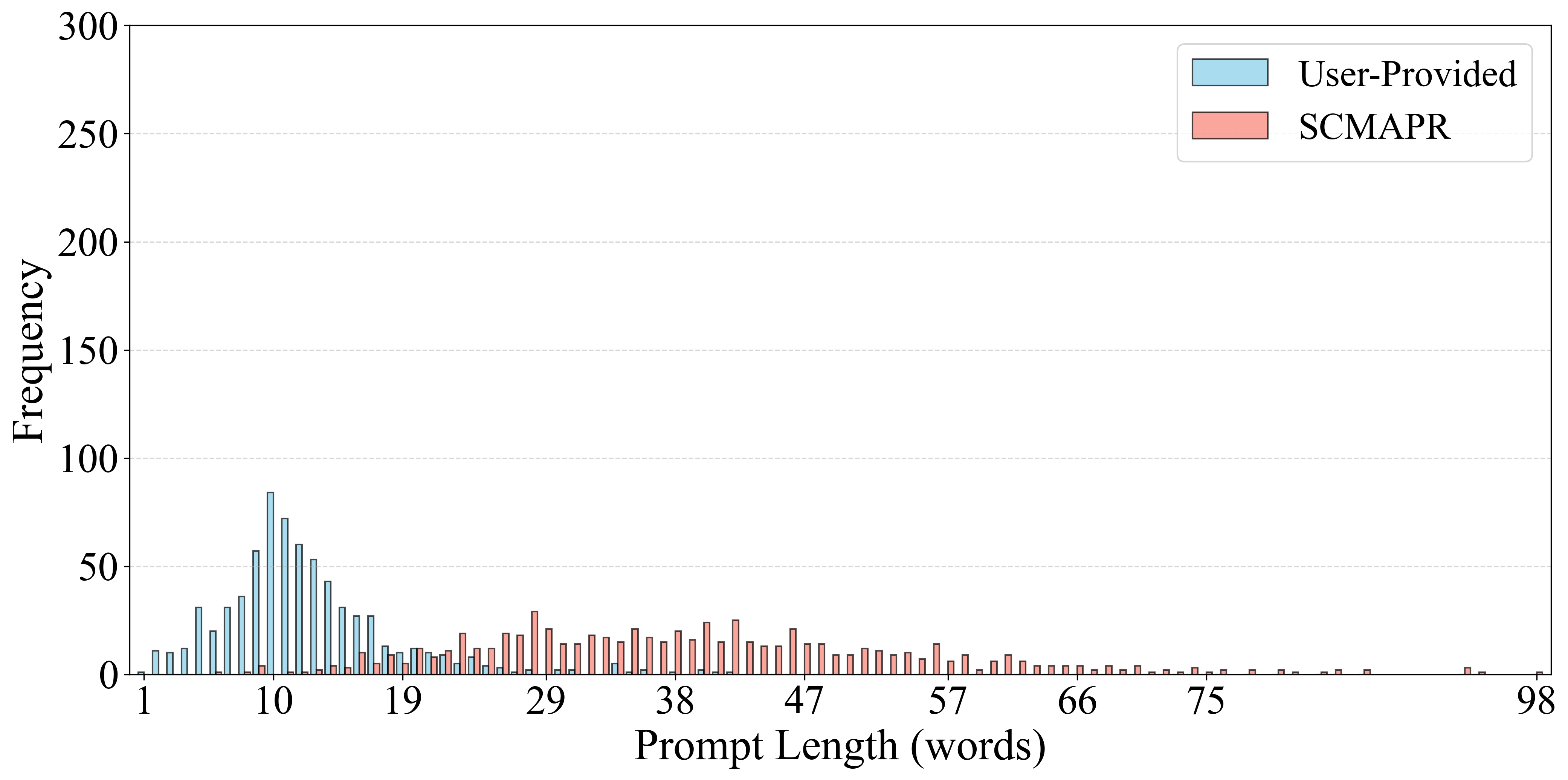}
    \caption{EvalCrafter}
  \end{subfigure}
  \caption{Distribution of prompt lengths on VBench and EvalCrafter (in words).}
  \label{fig:statistics_prompt_length_1}
\end{figure*}

\begin{figure*}[!h]
  \centering
  \captionsetup[subfigure]{justification=centering}
  \begin{subfigure}[b]{0.7\textwidth}
    \centering
    \includegraphics[width=\linewidth]{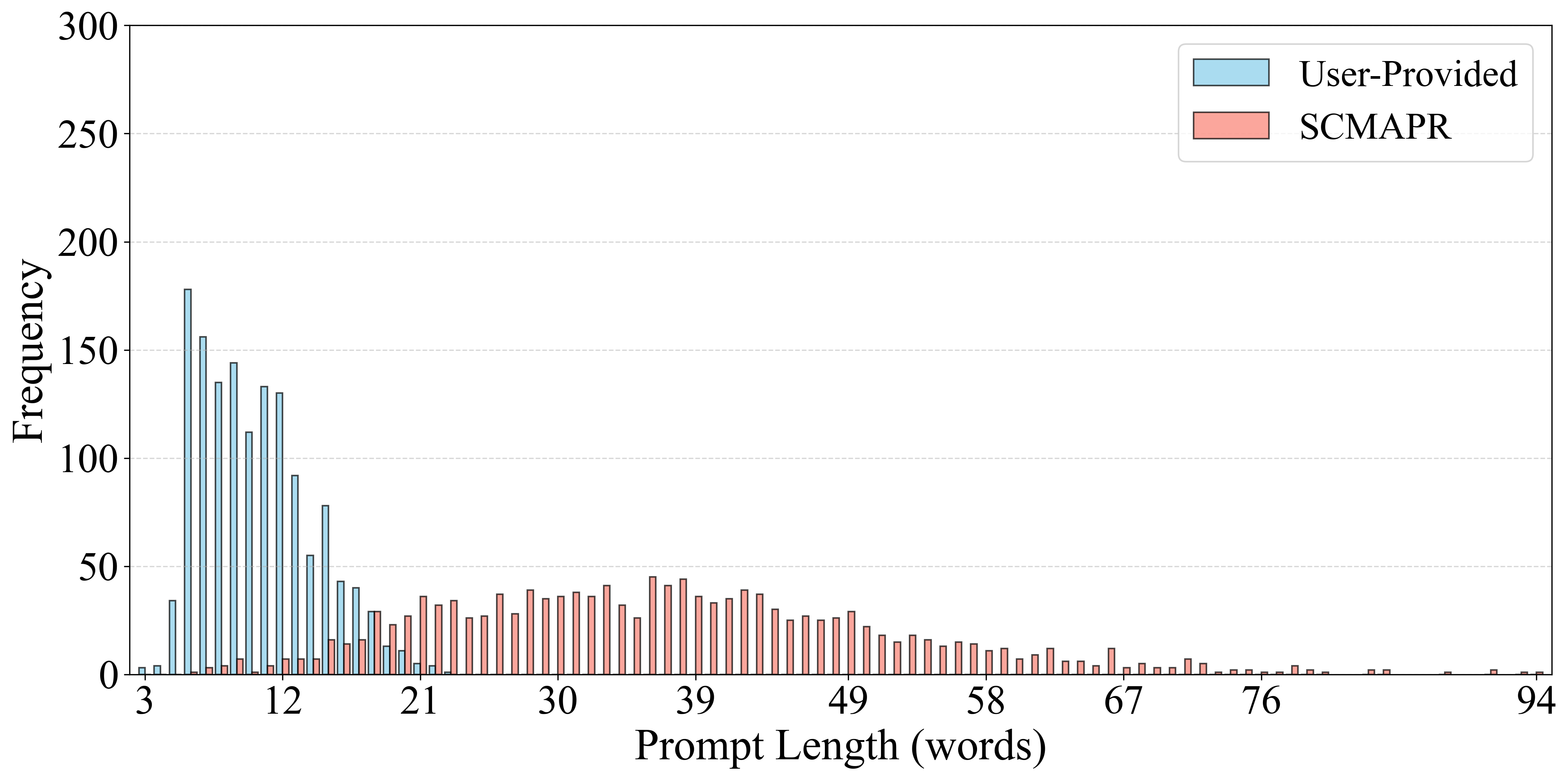}
    \caption{T2V-CompBench}
  \end{subfigure}

\begin{subfigure}[b]{0.7\textwidth}
    \centering
    \includegraphics[width=\linewidth]{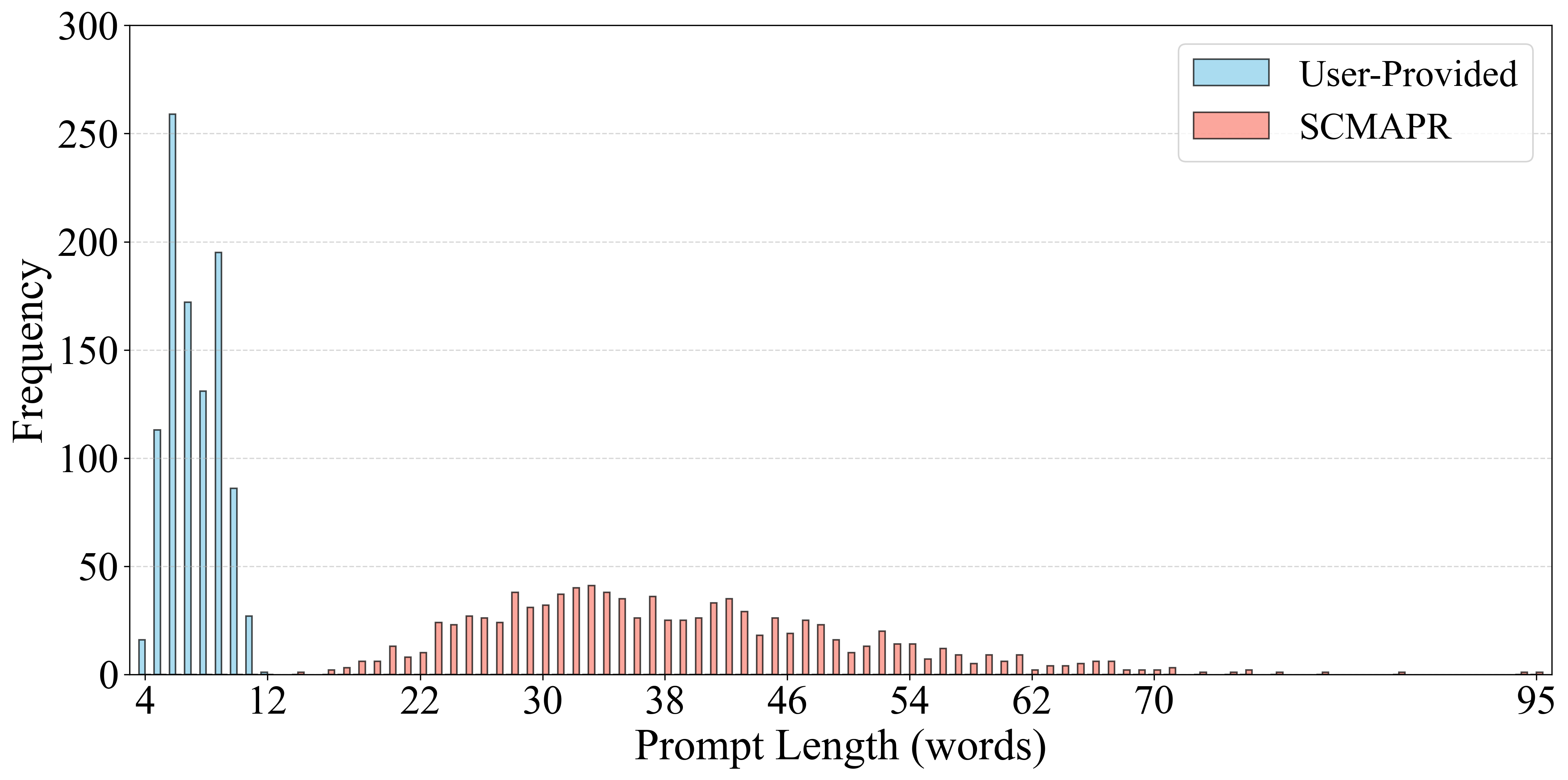}
    \caption{T2V-Complexity}
  \end{subfigure}
  \caption{Distribution of prompt lengths on T2V-CompBench and T2V-Complexity (in words).}
  \label{fig:statistics_prompt_length_2}
\end{figure*}

\section{Additional Experiments}
In this section, we present additional experiments, including qualitative examples for supplementary analysis of scenario distribution and prompt length that motivate T2V-Complexity.

\subsection{Analysis on Scenario Distribution}\label{supp-subsec:analysis_on_scenario_distribution}
Figures~\ref{supp-fig:distribution_vbench}, \ref{supp-fig:distribution_evalcrafter}, and \ref{supp-fig:distribution_compbench} present the scenario distributions of user-provided prompts in VBench, EvalCrafter, and T2V-CompBench. Collectively, the three benchmarks contain 946, 700, and 1,400 prompts.

In VBench, we observe that 32.9\% of the prompts fall into complex scenario categories, while the remaining 67.1\% are classified as non-complex. In EvalCrafter, complex scenarios account for 60.3\% of all prompts, with 39.7\% categorized as non-complex. T2V-CompBench exhibits an even higher proportion of complex scenarios, reaching 69.4\%, compared to 30.6\% non-complex prompts. These statistics indicate that complex scenarios constitute a substantial fraction of prompts encountered in practical T2V generation benchmarks.

However, the complex-scenario categories exhibit severe imbalance across all three benchmarks. For example, Multi-Element Scenes are scarce in VBench, while Abstract Descriptions are nearly absent in EvalCrafter. Notably, Scene Transition does not appear in either VBench or EvalCrafter, and is only sparsely represented in T2V-CompBench. Similar long-tail phenomena are observed for other categories, including Stylistic Hybrids and Abstract Descriptions. 

Motivated by these observations, we introduce \textbf{T2V-Complexity}, a balanced benchmark explicitly designed to address category imbalance in complex scenarios. T2V-Complexity covers ten distinct complex scenario categories, each containing 100 curated prompts. This design enables systematic and controlled evaluation of T2V models across diverse forms of prompt complexity that are often overlooked or insufficiently covered in existing benchmarks.

\subsection{Analysis on Prompt Length}
Figures~\ref{fig:statistics_prompt_length_1} and~\ref{fig:statistics_prompt_length_2} report word-level prompt length distributions of user-provided prompts (user inputs) and the refined prompts produced by \name{} on four benchmarks, including VBench (946 prompts), EvalCrafter (700 prompts), T2V-CompBench (1400 prompts), and T2V-Complexity (1000 prompts). Across all benchmarks, user-provided prompts are always short and heavily concentrated in the low-word interval, which limits their ability to specify details such as fine-grained visual attributes, spatial layouts, or temporal relations. This limitation is more pronounced in complex scenarios, where the generation target often depends on additional constraints that are rarely stated in brief user-provided prompts.

After refinement, the length distributions consistently shift toward larger values across all benchmarks. On VBench, user-provided prompts are mainly concentrated within approximately 1 to 12 words, while refined prompts extend to a substantially broader range and reach about 82 words.
On EvalCrafter, user-provided prompts cluster around 8 to 15 words, while refined prompts form a long tail that reaches about 98 words.
On T2V-CompBench, user-provided prompts are mostly within roughly 3 to 20 words, while refined prompts span a wider range and reach about 94 words.
On T2V-Complexity, user-provided prompts are tightly concentrated within roughly 4 to 12 words, while refined prompts consistently fall in a substantially longer interval and extend to about 95 words.

More importantly, this increase in prompt length is not due to indiscriminate extension of sentences.
Instead, \name{} enriches the prompts by explicating implicit requirements and adding scenario-relevant details while preserving user intent.
As a result, the refined prompts provide more explicit and actionable specifications for downstream text-to-video generation.

\subsection{Analysis on Efficiency of Conditional Revision }
To validate the efficiency of conditional revision, we provide an additional analysis from three perspectives: the iteration frequency on two benchmarks, the effectiveness of additional rounds in improving semantic fidelity, and the behavior of diminishing returns as the number of rounds increases.

\begin{table}[t]
\centering
\caption{Statistics of rewrite rounds on VBench.}
\label{tab:vbench_rewrite_statistics}
\begin{tabular}{lcc}
\toprule
VBench & Quantity & Proportion \\
\midrule
rewrite once    & 818 & 86.47\% \\
rewrite twice   & 118 & 12.47\% \\
rewrite 3 times & 10  & 1.06\%  \\
Total           & 946 & 100.00\% \\
\bottomrule
\end{tabular}
\end{table}

\begin{table}[t]
\centering
\caption{Statistics of rewrite rounds on T2V-Complexity.}
\label{tab:t2v_complexity_rewrite_statistics}
\begin{tabular}{lcc}
\toprule
T2V-Complexity & Quantity & Proportion \\
\midrule
rewrite once    & 823  & 82.30\%  \\
rewrite twice   & 151  & 15.10\%  \\
rewrite 3 times & 24   & 2.40\%   \\
rewrite 4 times & 2    & 0.20\%   \\
Total           & 1000 & 100.00\% \\
\bottomrule
\end{tabular}
\end{table}

\textbf{Iteration frequency.} We conduct an additional analysis without limiting the maximum number of revision rounds and study the iteration behavior on both VBench and T2V-Complexity. The results are shown in Table~\ref{tab:vbench_rewrite_statistics} and Table~\ref{tab:t2v_complexity_rewrite_statistics}, respectively. 
SCMAPR enforces a strict acceptance criterion, accepting a refined prompt only when semantic coverage reaches $p_{ET}=1$ and the contradiction rate is $p_{CT}=0$. Otherwise, SCMAPR triggers conditional revision until the criterion is met. On VBench, 86.47\% (818/946) of prompts are verified after one rewrite, 12.47\% (118/946) require two rewrites, and 1.06\% (10/946) require three rewrites. On T2V-Complexity, 82.30\% (823/1000) converge in one rewrite, while 15.10\% (151/1000), 2.40\% (24/1000), and 0.20\% (2/1000) require two, three, and four rewrites, respectively.

\begin{table*}[t]
\begin{center}
\begin{minipage}{0.93\linewidth}
\raggedright
\textbf{User Input}: A dim glow appearing from nowhere.

\textbf{Refined Prompt}: In the quiet stillness of a dark, misty forest, a soft, pulsating blue glow emerges from a crack in an ancient, moss-covered tree trunk. The light gradually intensifies, casting eerie, shifting shadows across the damp ground and gnarled roots.
\end{minipage}
\end{center}
\vspace{2mm}

\centering
\noindent \textbf{Verification Results (Refined Prompt)}
\begin{tabular}{c c c c l}
\toprule
\textbf{i} & \textbf{Category} & \textbf{Atom} & \textbf{Label} & \multicolumn{1}{c}{\textbf{Reason}} \\
\midrule
0 & objects   & glow      & ET & Evidence explicitly mentions `soft, pulsating blue glow'. \\
1 & actions   & appearing & MS & No mention of any entity appearing. \\
2 & locations & nowhere   & MS & Evidence describes a specific location, not `nowhere'. \\
3 & scenery   & dim       & MS & Evidence describes intensifying light, not dimness. \\
\bottomrule
\end{tabular}
\noindent Coverage: 0.250 \quad Contradiction: 0.000
\caption{Verification Results for first prompt refinement.}
\label{tab:verification_report_llm_entailment_1}
\end{table*}
\begin{table*}[t]
\begin{center}
\begin{minipage}{0.93\linewidth}
\raggedright
\textbf{Revised Prompt}: In the quiet stillness of a dark, misty forest, a soft, pulsating blue glow emerges from a crack in an ancient, moss-covered tree trunk. The light \textbf{emerges slowly}, casting eerie, shifting shadows across the damp ground and gnarled roots.
\end{minipage}
\end{center}
\vspace{2mm}

\centering
\noindent \textbf{Verification Results (Revised Prompt)}
\begin{tabular}{c c c c p{8.5cm}}
\toprule
\textbf{i} & \textbf{Category} & \textbf{Atom} & \textbf{Label} & \multicolumn{1}{c}{\textbf{Reason}} \\
\midrule
0 & objects   & glow      & ET & Evidence explicitly mentions ``soft, pulsating blue glow''. \\
1 & actions   & appearing & ET & The light emerges slowly, which implies appearing. \\
2 & locations & nowhere   & MS & Atom ``nowhere'' not mentioned or implied in evidence. \\
3 & scenery   & dim       & ET & Dark forest and shadows imply dim lighting. \\
\bottomrule
\end{tabular}
\noindent Coverage: 0.750 \quad Contradiction: 0.000
\caption{Verification Results for second prompt refinement (first revision).}
\label{tab:verification_report_llm_entailment_2}
\end{table*}

\begin{table*}[t]
\begin{center}
\begin{minipage}{0.93\linewidth}
\raggedright
\textbf{Revised Prompt}: In the quiet stillness of a dark, misty forest, a soft, pulsating blue glow \textbf{appears as if from nowhere}, with \textbf{no visible source}. The light emerges slowly, casting eerie, shifting shadows across the damp ground and gnarled roots.
\end{minipage}
\end{center}
\vspace{2mm}

\centering
\noindent \textbf{Verification Results (Revised Prompt)}
\begin{tabular}{c c c c p{8.5cm}}
\toprule
\textbf{i} & \textbf{Category} & \textbf{Atom} & \textbf{Label} & \multicolumn{1}{c}{\textbf{Reason}} \\
\midrule
0 & objects   & glow      & ET & Evidence explicitly mentions ``soft, pulsating blue glow''. \\
1 & actions   & appearing & ET & The light emerges slowly, which implies appearing. \\
2 & locations & nowhere   & ET & Refined prompt explicitly states the glow appears ``as if from nowhere''. \\
3 & scenery   & dim       & ET & Dark forest and shadows imply dim lighting. \\
\bottomrule
\end{tabular}
\noindent Coverage: 1.000 \quad Contradiction: 0.000
\caption{Verification Results for third prompt refinement (second revision).}
\label{tab:verification_report_llm_entailment_3}
\end{table*}

\textbf{Effectiveness of additional rounds.} Importantly, additional revision rounds are not redundant iterations. Instead, they perform targeted corrections guided by atom-level verification feedback. Each triggered round is driven by previously detected missing or contradictory atoms and refines the prompt toward full semantic coverage under a fixed atom set. We provide a representative three-round (two revisions) case in Table~\ref{tab:verification_report_llm_entailment_1}, \ref{tab:verification_report_llm_entailment_2} and \ref{tab:verification_report_llm_entailment_3} for clarity and note that similar examples are also included in Section~\ref{supp-sec:qualitative_examples_and_case_study}.

Starting from the user input that “A dim glow appearing from nowhere”, the first refined prompt over-specifies a concrete source, namely a crack in a tree trunk. It also describes an intensifying light. As a result, three atoms are marked as missing, including appearing, nowhere, and dim. The coverage is 0.25 and the contradiction rate is 0.00.

In the first revision round, SCMAPR makes the appearance process explicit by stating that the glow emerges slowly. This resolves the missing action atom and increases coverage to 0.75. The contradiction rate remains 0.00.

In the second revision round, SCMAPR explicitly indicates an ungrounded source by stating that the glow appears as if from nowhere and has no visible source. This resolves the remaining missing atom and achieves full semantic coverage. The final coverage reaches 1.00 with a contradiction rate of 0.00. This case study shows that additional rounds provide measurable gains in semantic fidelity until the acceptance criterion is satisfied.

\textbf{Diminishing returns.} We observe a clear pattern of diminishing returns beyond the second iteration. A second revision round is required only for a minority of prompts, accounting for 12.47\% on VBench and 15.10\% on T2V-Complexity. The proportion of prompts that require further rounds then drops sharply. On VBench, only 1.06\% require three rounds. On T2V-Complexity, at most 2.60\% require three or four rounds. Nevertheless, these additional rounds remain beneficial for the hard tail. They continue to improve semantic fidelity until the strict acceptance criterion is satisfied, rather than degrading prompt quality.


\textbf{Setting the maximum round.} Based on these observations, we set the maximum number of revision rounds to five in our experiments. This budget covers the long-tail cases requiring multiple revisions while avoiding unnecessary repeated LLM calls.

\section{Instructions}
\label{app-sec:router_prompt}
In this section, we provide prompt templates for multiple agents employed to implement \name{}.

\lstset{
    basicstyle=\ttfamily\small,
    breaklines=true,
    keywordstyle=\color{blue},
    stringstyle=\color{red},
    commentstyle=\color{green!50!black},
    showstringspaces=false,
    literate={*}{{*}}1
             {≤}{{$\le$}}1
             {≥}{{$\ge$}}1
             {`}{\texttt{\char96}}1
             {<}{{\textless}}1
             {>}{{\textgreater}}1
             {\{}{{\{}}1
             {\}}{{\}}}1
}

\subsection{Prompt for Scenario Tagging (Scenario Router)}\label{app-subsec:prompt_for_scenario_tagging}
\begin{lstlisting}
You are a Scenario Router Agent for Text-to-Video (T2V) prompt refinement.

Return ONLY a valid JSON object of the exact form:
{{"label": "<one of SCENARIO_TAGS>", "reason": "<short phrase (<= 20 words)>"}}

SCENARIO_TAGS (choose exactly one):
1) Abstract Descriptions
2) Complex Spatial Relations
3) Multi-Element Scenes
4) Fine-Grained Appearance
5) Temporal Consistency
6) Stylistic Hybrids
7) Causality & Physics
8) Camera Motion
9) Object Interaction
10) Scene Transitions
11) Non-difficult

## Task
Given a short English prompt P_in, decide which SINGLE tag best describes the dominant difficulty a T2V model would face when generating a video.

## Diagnostic definitions (brief)
- Abstract Descriptions: metaphorical/symbolic/abstract intent; requires semantic grounding beyond literal objects.
- Complex Spatial Relations: explicit geometric relations (left/right/between/center/above/behind) that must be satisfied.
- Multi-Element Scenes: high visual density; many salient entities/objects; preserving completeness and counts.
- Fine-Grained Appearance: identity/textures/text/small details/materials are essential.
- Temporal Consistency: time evolution or long-range continuity is central (blooming, melting, state progression).
- Stylistic Hybrids: multiple distinct styles must co-exist coherently (e.g., oil painting + cyberpunk).
- Causality & Physics: cause-effect chains or physically plausible dynamics are required (falling, shattering, splashing).
- Camera Motion: camera trajectory is central (pan/tilt/zoom/orbit/tracking shot).
- Object Interaction: explicit contact/manipulation between entities (pick up/pour/collide/grasp), interaction-driven motion/occlusion.
- Scene Transitions: multi-shot structure, cuts, or transitions are essential.
- non-difficult: none of the above applies.

## Tie-breaking priority (when multiple apply)
1) Abstract intent dominates -> Abstract Descriptions
2) Explicit spatial constraints dominate -> Complex Spatial Relations
3) Many entities / dense scene dominates -> Multi-Element Scenes
4) Fine-grained/identity/textural constraints dominate -> Fine-Grained Appearance
5) Temporal evolution / continuity dominates -> Temporal Consistency
6) Style blending dominates -> Stylistic Hybrids
7) Cause-effect / physical plausibility dominates -> Causality & Physics
8) Camera trajectory dominates -> Camera Motion
9) Contact-driven interaction dominates -> Object Interaction
10) Multi-shot transitions dominates -> Scene Transitions
11) Otherwise choose -> non-difficult

## Few-Shot Examples
- "Hope dances in a field of forgotten dreams."
  -> Abstract Descriptions

- "A cat sits between a dog and a parrot hovering above them."
  -> Complex Spatial Relations

- "Ten performers dance under fireworks in a crowded plaza."
  -> Multi-Element Scenes

- "A close-up of a cracked porcelain cup with visible glaze texture."
  -> Fine-Grained Appearance

- "A flower bud slowly opens into full bloom."
  -> Temporal Consistency

- "A medieval castle rendered in cyberpunk neon style."
  -> Stylistic Hybrids

- "A glass is pushed off a table and shatters on the floor."
  -> Causality & Physics

- "The camera slowly pans across a busy marketplace."
  -> Camera Motion

- "A person pours water into a cup and places it down."
  -> Object Interaction

- "The scene cuts from a city street to a quiet bedroom at night."
  -> Scene Transitions

- "A child runs across a field." -> non-difficult

Classify the following prompt:
P_in: {P_in}
\end{lstlisting}

\subsection{Prompt for Policy Generation (Policy Generator)}
\label{app:policy_prompts}

As described in Section~\ref{subsec:stage_policy}, \name{} does not rely on fixed, category-specific meta-prompts.
Instead, a policy agent synthesizes a prompt-specific rewriting policy $\pi$ conditioned on the user input $P_{\text{user}}$ and the routed scenario tag $\hat{y}$.
This appendix provides the instruction prompts, output schema, and representative policy exemplars employed to implement Stage II--III, together with the structured feedback interface for conditional revision (Stage V).

\begin{lstlisting}
You are a policy generator for a text-to-video prompt refinement system. 

Your task is to generate a policy that reshapes user inputs into concise intents, principles and rules for video generation, ensuring that no new facts are introduced while maintaining fidelity to the original meaning.

You will be given:
(1) A user input P_user for text-to-video generation
(2) A routed scenario tag y_hat and its definition

This system focuses on scenario-aware prompt rewriting.
The routed tag y_hat may indicate either a non-difficult case or one of the following 10 complex scenario categories including:
Abstract Descriptions; Complex Spatial Relations; Multi-Element Scenes; Fine-Grained Appearance;
Temporal Consistency; Stylistic Hybrids; Causality & Physics; Camera Motion;
Object Interaction; Scene Transitions.

Your task: 
synthesize a prompt-specific rewriting policy pi to guide a downstream Prompt Refiner.
The policy should be conditioned on y_hat and reflect the main challenges described in its definition.
When y_hat corresponds to a non-difficult case, the policy should remain minimal and conservative.

Guidelines:
- Scenario conditioning: Shape the policy with respect to y_hat and its scenario-specific guidance.
- Fidelity: Preserve the intended meaning and stated content.
- Practicality: Express the policy as clear, executable guidance for prompt rewriting.

Scenario-specific guidance (use y_hat to decide emphasis):
- Abstract Descriptions: 
(1) **Clarify abstract imagery or concept:** Abstract imagery or concept may be concretized through grounded, naturalistic entities, scenes, or atmosphere that express the intended meaning. 
(2) **Creative instantiation**: Creative instantiation is strongly encouraged if it serves the abstraction and enhances the emotional or conceptual depth of the scene.
(3) **Avoid ambiguous instances**: Specify characters, entities, objects or scenes clearly and avoid using vague, ambiguous, or unclear terms that may cause confusion or multiple interpretations.
  For example, do not use "human-like" or "human", because they are vague. Instead, use terms like "girl", "boy", "young woman" or "young man" as they are more specific.
(4) **Ensure adherence to theme**: The generated scene should align with the specified theme, particularly the adjectives used in the description. Avoid introducing unrelated elements that deviate from the intended atmosphere or message of the scene.

- Complex Spatial Relations: 
(1) **Emphasize spatial clarity:** Explicitly describe positions, distances, and relative orientations of elements in the scene.
(2) **Position characters by relationship:** Place adversarial characters on opposite sides. Place non-adversarial characters between the adversarial characters.
(3) **Assign appropriate actions**: Define suitable and clear movements or actions for each character.
(4) **Maintain key details:** Preserve all essential objects, actions, characters, and environments.
    
- Multi-Element Scenes:
(1) **Preserve all key elements**: Keep essential characters, objects, settings, and relationships.
(2) **Simplify structure**: Avoid unnecessary adjectives or complex phrasing.
(3) **Ensure temporal and spatial clarity**: Present events in a logical and visually coherent order.

- Fine-Grained Appearance: 
(1) **Preserve Fine-Grained Details**: Keep all essential visual attributes (colors, textures, facial expressions, clothing, environmental elements, etc.) while removing irrelevant or repetitive details.
(2) **Enhance Visual Clarity**: Use precise and descriptive language to clearly define characters, objects, actions, and spatial relationships, making the scene easy for the model to interpret.
(3) **Add Cinematic Guidance**: Optionally introduce cinematic elements like lighting, camera movement, focus depth, or shot composition to improve video realism.
(4) **Maintain Logical Structure**: Ensure actions and events are described in chronological order with clear transitions, avoiding ambiguity or contradictions.
(5) **Optimize for Video Generation**: Emphasize motion cues, scene continuity, and environmental context so the model can generate smooth, coherent multi-frame sequences.

- Temporal Consistency:
(1) **Be Clear and Explicit:** Turn ambiguous or compressed descriptions into precise phrases.
(2) **Be Scene-Oriented:** Clearly separate and describe characters, objects, locations, and actions.
(3) **Follow Logical Order:** Present elements in a clear sequence (foreground -> background; primary -> secondary; chronological actions).
(4) **Preserve All Key Details:** Keep every important visual detail while removing redundancies.
(5) **Include Style and Lighting:** Explicitly state any implied visual style, palette, or lighting.
    
- Stylistic Hybrids:
(1) **Scene Composition:** Specify key subjects, actions, and environments in short, direct phrases.
(2) **Visual Consistency:** Resolve ambiguity about style blending or scene layout.
(3) **Compactness:** Use minimal yet descriptive language; no filler words.
    
- Causality & Physics:
(1) **Preserve Meaning:** Retain all key entities, actions, and causal relationships.
(2) **Physics Clarity:** Clearly state motion, timing, and forces.
(3) **Morphological Changes:** Emphasize transformations in object shape, size, or state over time.
(4) **Logical Flow:** Present actions in chronological order.

- Camera Motion:
(1) Be Clear on Movement: Specify camera movement only when stated or clearly implied by the user input (e.g., "slowly pans across the marketplace").
(2) Smooth Transitions: Use smooth and continuous camera movement unless abrupt motion is specified.
(3) Follow the Action: Ensure camera movements follow the flow of action or emphasize key moments in the scene.
(4) Maintain Stability when No Movement is Implied: If no camera movement is suggested, keep the camera fixed to avoid distracting the viewer.
(5) Enhance Mood and Emphasis: Camera movements should reinforce the emotional tone and emphasis of the scene (e.g., zooming in for a close-up or panning for a panoramic view).
    
- Object Interaction:
(1) Define Clear Interactions: Ensure interactions between objects or characters are depicted clearly, specifying actions (e.g., "pours water into a cup").
(2) Respect Cause and Effect: Ensure that the actions and reactions between objects or characters are logically consistent (e.g., "pushed off the table and shattered").
(3) Focus on Interaction Dynamics: Show the dynamics of the interaction, such as force, direction, and timing (e.g., the glass breaking upon impact).
(4) Keep Spatial Consistency: Ensure that interactions respect the spatial relationships described in the prompt (e.g., cup placed on the table, liquid spilling, etc.).
(5) Avoid Over-complication: Keep interactions simple and avoid introducing unnecessary complexity unless explicitly required.

- Scene Transitions:
(1) Ensure Smooth Transitions: Ensure that transitions between scenes are seamless, with clear visual cues or shifts in time, space, or mood.
(2) Clarify Context Shifts: If transitioning from one location to another (e.g., city street to bedroom), ensure the viewer understands the change through visual cues like lighting, architecture, or props.
(3) Maintain Continuity: Ensure that essential elements from the previous scene are carried over or referenced in the transition to maintain continuity (e.g., a person leaving a room and entering a new one).
(4) Emphasize Emotional Shift: Use transitions to underline the emotional shift, if any, between scenes (e.g., a sudden contrast from a bustling street to a calm, quiet bedroom).
(5) Use Cinematic Techniques: If applicable, use cinematic techniques like fades, dissolves, or cuts to highlight the transition without disrupting the flow of the story.
    
- Non-Difficulty:
(1) **Improve Clarity:** Rewrite in clear, simple language to eliminate ambiguity or vagueness.
(2) **Model-Friendly Syntax:** Ensure the prompt is straightforward for machine interpretation and avoid figurative language or unnecessary modifiers.
(3) **Direct Scene Description:** Describe the scene plainly, focusing only on necessary visual elements.


Return STRICT JSON with keys:
- "policy": an OBJECT with keys:
    - "intent": 1-2 sentences describing the intent of user query according to user input.
    - "principles": 1-3 sentences encouraging prompt refiner to rewrite detailed prompts.
    - "rules": 2-6 sentences describing executable Scenario-specific guidelines.
No other keys.

P_user:
{p_user}

y_hat:
{y_hat}

Definition:
{y_def}
\end{lstlisting}

\subsection{Prompt for Policy-Conditioned Refinement (Prompt Refiner)}
\label{subsec:refine}

\begin{lstlisting}
You are a **Prompt Refiner** designed to refine user inputs for text-to-video generation. Your task is to rewrite the user input to make it clearer, more detailed, and suitable for generating high-quality video content, while using the provided policy as a reference.

## Task
**Refining User input Based on Policy**

## Role
As a Prompt Refiner, your main responsibility is to take the user input and enhance it. The goal is to ensure the prompt is:
   - Clear and well-defined, with key details emphasized.
   - Concise yet descriptive, conveying all necessary information for accurate video generation.
   - Aligned with the general **intent** of the policy, with reference to the **principles** and **rules** as helpful guidance.

## Refinement Objectives
   - Preserve the original meaning and intent of the user input.
   - Make the prompt more detailed by expanding on the key elements (such as characters, objects, actions, and settings) where necessary.
   - Avoid introducing new concepts or elements unless they are implied or necessary for a more complete description.
   - Ensure that the refined prompt maintains a consistent tone, mood, and atmosphere based on the user intent.

## Task-Specific Instructions
   - **Clarify key details**: Expand on vague or under-specified parts of the prompt. For example, if the prompt mentions a "beautiful sunset", you might clarify what makes it beautiful (e.g., warm tones, fading light).
   - **Keep it concise yet informative**: Ensure that the prompt is not overly verbose but still provides enough detail to accurately generate the video. Avoid redundancy and unnecessary restatements.
   - **Ensure emotional and thematic consistency**: While refining the prompt, make sure the tone matches the intent, whether it is serene, exciting, melancholic, etc.

## Example Policy for Refinement
   - **Intent**: Preserve the serene, monumental desert landscape described in the user input. Ensure the generated video depicts a single, massive sandstone arch dominating a tranquil Utah desert scene.
   - **Principles**: 
     - Maintain the specific geographical and geological setting.
     - Emphasize stillness, scale, and natural beauty.
   - **Rules**:
     - The scene must be set in the Utah desert. Do not change the location.
     - The central subject is a single, massive sandstone arch. It must be the dominant visual element.
     - The arch must span the horizon, implying great width and a low, panoramic perspective.
     - The overall mood must be tranquil and still. Avoid any dynamic action, weather, or human/animal presence.
     - The lighting and color palette should reflect a natural desert environment (e.g., warm tones, clear sky).
     - The camera should be stable, with a wide, establishing shot that captures the full span of the arch against the horizon.

## Output Requirements
   - **Output only the rewritten prompt**: The refined prompt should be returned in multiple concise sentences, with no explanations or extraneous content.
   - Do not introduce new elements or actions that were not present in the original prompt unless they are implied by the policy or necessary for clarity.
   - Ensure that the rewritten prompt aligns with the core intent and tone, while being as clear and descriptive as possible.

USER INPUT:
{user_input}

POLICY:
Intent:
{intent}

Principles:
{principles}

Rules:
{rules}
\end{lstlisting}

\subsection{Prompt for Atomization (Semantic Atomizer)}
\label{subsec:App:Atomization}

\begin{lstlisting}
You are an information extractor.
    
## Strict Constraints:
1) Only output atoms that appear verbatim in the given prompt (exact surface spans).
2) Do NOT paraphrase, generalize, translate, lemmatize, or infer missing items.
3) Each atom must be a substring of the prompt. If you cannot find it exactly, do NOT output it.
4) Keep the original casing and wording as in the prompt.
5) Output ONLY valid JSON with keys: characters, objects, actions, locations, scenery.
6) If an abstract concept is explicitly used as an entity/actor in the prompt (e.g., "Hope", "Time", "Love"), it is allowed to be included in atoms list (see example 2).
7) Each list item is 1-4 words copied from the prompt (no extra punctuation).

## Example 1
User input:
A cat plays chess with a dog while a parrot referees in a steampunk library.

Output:
{
  "characters": ["cat", "dog", "parrot"],
  "objects": ["chess"],
  "actions": ["plays", "referees"],
  "locations": ["library"],
  "scenery": ["steampunk"]
}

## Example 2
User input:
Hope drifting somewhere far away.

Output:
{
  "characters": ["Hope"],
  "objects": [],
  "actions": ["drifting"],
  "locations": ["somewhere far away"],
  "scenery": []
}

Now extract from the user input.
Return JSON only. No commentary.
\end{lstlisting}

\begin{figure*}[t]
    \centering 
    \includegraphics[width=\textwidth]{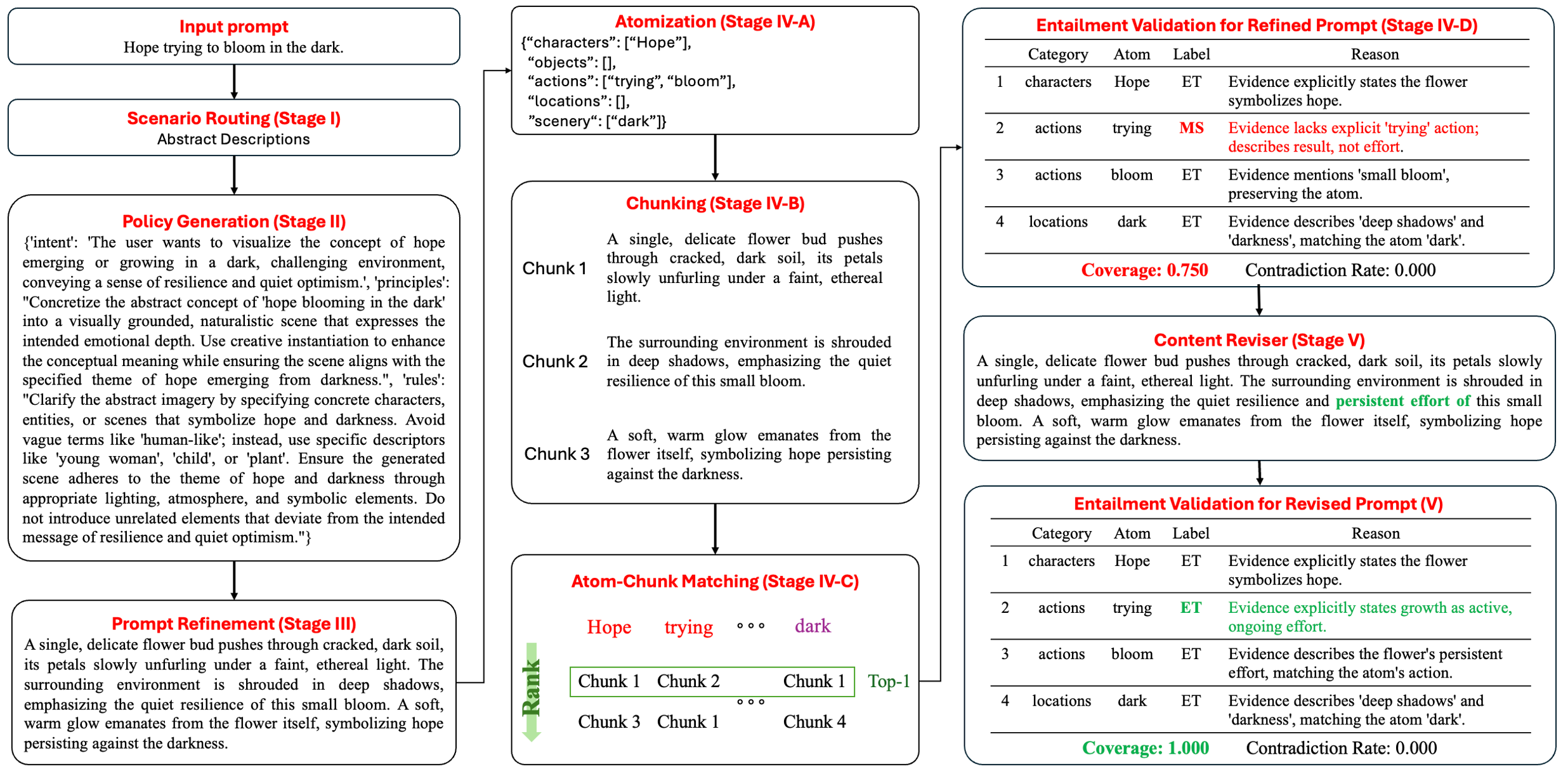}
    \caption[Illustration of Semantic Verification]{
        An end-to-end case study of \name{} with self-correction. Given a user input, the framework performs scenario routing, policy generation, policy-conditioned prompt refinement, atom-level verification, and targeted revision. Entailment Validator labels each atom-evidence pair and conditionally triggers targeted revision, producing a verified refined prompt for downstream video generation.
    }
    \label{fig:case_study_1} 
\end{figure*}

\subsection{Prompt for Validation (Entailment Validator)}
\label{subsec:App:entailment}

\begin{lstlisting}
You are an Entailment Validator for a prompt-refinement system.

## Task
Given an atom (a minimal semantic constraint) extracted from the ORIGINAL prompt, 
and evidence text from the REFINED prompt, decide the relation:
- ET (entailment): the refined prompt clearly preserves the atom.
- MS (missing): the refined prompt does not state/support the atom.
- CT (contradiction): the refined prompt states something incompatible with the atom.

## Output Format
Return ONLY a JSON object in the exact format:
'{"label": "ET|MS|CT", "reason": "<= 25 words"}'

## Rules
- Use only the provided evidence + refined prompt (if included).
- If the evidence is insufficient to confirm the atom, choose MS.
- CT only if there is explicit conflict (negation, different entity/count, incompatible attribute).
- Do not add any extra keys.
\end{lstlisting}

\subsection{Prompt for Revision (Content Reviser)}
\begin{lstlisting}
You are a prompt revision agent.

## Input
You will be given:
- an original prompt (ground-truth intent)
- a refined prompt (possibly flawed)
- a verification report listing atomic constraints labeled as:
  ET (entailed), MS (missing), CT (contradiction)

## Task
1) Fix ALL MS constraints by making them explicit in the refined prompt.
2) Fix ALL CT issues by removing or rewriting conflicting statements in the refined prompt.
3) Preserve everything in the refined prompt that does NOT conflict with the original prompt.
4) Do NOT add new facts/entities not present in the original prompt.
5) Apply minimal edits. Prefer adding a compact 'Constraints': block at the end for MS.
6) For CT, prefer deleting or rewriting the conflicting phrases; the original prompt has priority.

## Output rules
- Output ONLY the revised prompt text.
- Do NOT output JSON unless asked.
- Do NOT add explanations.

ORIGINAL PROMPT:
{original_prompt}

CURRENT REFINED PROMPT:
{refined_prompt}

VERIFICATION ISSUES (MS/CT):
{json.dumps(payload, ensure_ascii=False, indent=2)}
\end{lstlisting}

\begin{table*}[ht]
\centering
\noindent \textbf{Verification Results (Refined Prompt)}
\begin{tabularx}{\textwidth}{ccccX}
\toprule
\textbf{i} & \textbf{Category} & \textbf{Atom} & \textbf{Label} & \multicolumn{1}{c}{\textbf{Reason}} \\ 
\midrule
0 & characters & Hope   & ET & Evidence explicitly states the flower symbolizes hope.  \\ 
1 & actions    & trying & MS & Evidence lacks explicit ``trying'' action; describes result, not effort. 
 \\ 
2 & actions  & bloom  & ET & Evidence mentions ``small bloom'', preserving the atom.  \\ 
3 & scenery    & dark & ET & Evidence describes ``deep shadows'' and ``darkness'', matching the atom ``dark''. \\ 
\bottomrule
\end{tabularx}
\noindent Coverage: 0.750 \hspace{1cm} Contradiction: 0.000
\caption{Validation results for refined prompt.}
\label{tab:verification_example_refined_prompt}
\end{table*}

\begin{table*}[ht]
\centering
\noindent \textbf{Verification Results (Revised Prompt)}
\begin{tabularx}{\textwidth}{ccccX}
\toprule
\textbf{i} & \textbf{Category} & \textbf{Atom} & \textbf{Label} & \multicolumn{1}{c}{\textbf{Reason}}\\ 
\midrule
0 & characters & Hope   & ET & Evidence explicitly states the flower symbolizes hope. 
 \\ 
1 & actions    & trying & ET & Evidence explicitly states growth as active, ongoing effort. 
 \\ 
2 & actions  & bloom  & ET & Evidence describes the flower's persistent effort, matching the atom's action.
 \\ 
3 & scenery    & dark & ET & Evidence describes ``deep shadows'' and ``darkness'', matching the atom ``dark''. \\ 
\bottomrule
\end{tabularx}
\noindent Coverage: 1.000 \hspace{1cm} Contradiction: 0.000
\caption{Validation results for revised prompt}
\label{tab:verification_example_revised_prompt}
\end{table*}

\begin{figure*}[ht]
    \centering 
    \includegraphics[width=\textwidth]{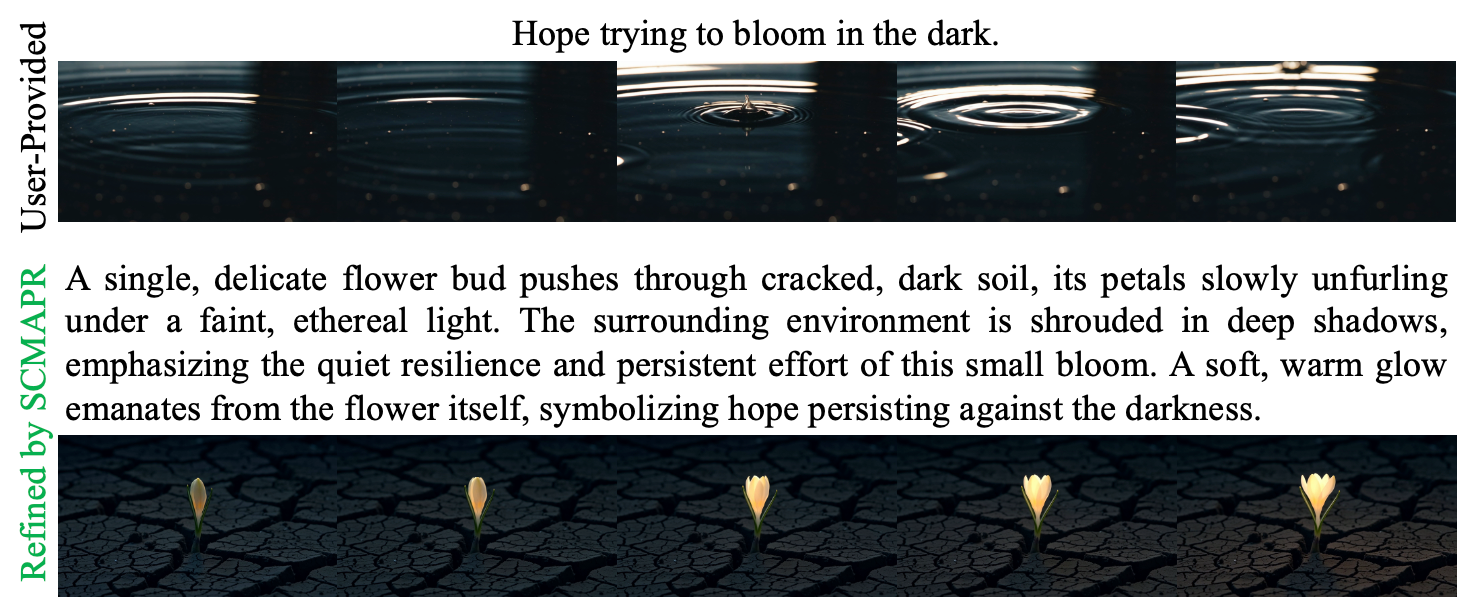}
    \caption{\textbf{Qualitative comparison of generated videos under an abstract user input (user-provided prompt) and refined prompt.}
    Sampled frames generated from the user-provided prompt (top) and the prompt refined by \name{} (bottom) are contrasted.
    The original prompt yields a visually plausible but semantically drifting dark-water scene, while the refined prompt concretizes the intended metaphor and produces a consistent depiction of a flower bud emerging through cracked dark soil with a warm glow, better matching the target concept.
    }
    \label{fig:case_study_video_results} 
\end{figure*}


\section{Qualitative Examples and Case Study}\label{supp-sec:qualitative_examples_and_case_study}
In this section, we provide representative examples that qualitatively illustrate the operation of \name{} and the intermediate artifacts produced by its stage-wise prompt refinement process.

\subsection{Example of Scenario-Conditioned Policy}
\label{app:policy_example}
Figure~\ref{fig:case_study_1} presents a complete end-to-end example of \name{}.
For readability, the following examples in this section are extracted from this figure.
We provide a representative example of a synthesized policy under the \textit{Abstract Descriptions} scenario.
This example demonstrates how Policy Generator concretizes an abstract user intent into visually grounded constraints and provides actionable rewriting rules.
\begin{lstlisting}
POLICY:
Intent:
The user wants to visualize the concept of hope emerging or growing in a dark, challenging environment, conveying a sense of resilience and quiet optimism.

Principles:
Concretize the abstract concept of 'hope blooming in the dark' into a visually grounded, naturalistic scene that expresses the intended emotional depth. Use creative instantiation to enhance the conceptual meaning while ensuring the scene aligns with the specified theme of hope emerging from darkness.

Rules:
Clarify the abstract imagery by specifying concrete characters, entities, or scenes that symbolize hope and darkness. Avoid vague terms like 'human-like'; instead, use specific descriptors like 'young woman', 'child', or 'plant'. Ensure the generated scene adheres to the theme of hope and darkness through appropriate lighting, atmosphere, and symbolic elements. Do not introduce unrelated elements that deviate from the intended message of resilience and quiet optimism.
\end{lstlisting}

\subsection{Illustrative Example of Refined and Revised Prompts}\label{supp-sub:example_of_refined_and_revised_prompts}
Section~\ref{supp-sub:example_of_refined_and_revised_prompts} presents a concrete case study to illustrate how revision complements prompt refinement in \name{}. We sequentially compare the original and revised prompts and their atom-level verification results before and after revision.


\subsubsection{Example of Revised Prompts Before and After Revision}
Content revision is designed to correct residual underspecification after prompt refinement by introducing minimal and targeted edits. In particular, it makes previously implicit intent explicit to reduce ambiguity and improve semantic fidelity. In the following example, we contrast the refined prompt with its revised counterpart produced by the Content Reviser. The corresponding verification results are reported in Section~\ref{supp-subsec:example_verification_results_before_and_after_revision}.

\textbf{Refined Prompt:} 
\textit{A single, delicate flower bud pushes through cracked, dark soil, its petals slowly unfurling under a faint, ethereal light. The surrounding environment is shrouded in deep shadows, emphasizing the quiet resilience of this small bloom. A soft, warm glow emanates from the flower itself, symbolizing hope persisting against the darkness.}

\textbf{Revised Prompt:} 
\textit{A single, delicate flower bud pushes through cracked, dark soil, its petals slowly unfurling under a faint, ethereal light. The surrounding environment is shrouded in deep shadows, emphasizing the quiet resilience and persistent effort of this small bloom. A soft, warm glow emanates from the flower itself, symbolizing hope persisting against the darkness.}

\subsubsection{Example of Verification Results Before and After Revision}\label{supp-subsec:example_verification_results_before_and_after_revision}
To highlight the effect of revision, we compare the atom-level verification results of the refined and revised prompts in Tables~\ref{tab:verification_example_refined_prompt} and \ref{tab:verification_example_revised_prompt}, respectively.

\textbf{Refined Prompt Verification (Table~\ref{tab:verification_example_refined_prompt}).}
Before revision, the verification results show a coverage of 75\%.
In particular, the action atom ``trying'' is labeled as missing, indicating that the refined prompt describes the outcome but does not explicitly convey effort. In the refined prompt, the scene is already grounded with concrete visual elements, but the intended notion of effort remains implicit. As reflected by the verification results in Table~\ref{tab:verification_example_refined_prompt}, the action atom “trying” is labeled as missing because the prompt primarily describes the outcome (a bloom in darkness) rather than an explicit attempt or ongoing struggle.

\textbf{Revised Prompt Verification (Table~\ref{tab:verification_example_revised_prompt}).} 
After revision, the coverage improves to 100\%.
By explicitly introducing ``persistent effort'', the revised prompt entails the atom ``trying'', strengthening semantic specificity for downstream T2V generation. In the revised prompt, we explicitly incorporate ``persistent effort'' to make the growth process active and intentional. This targeted revision turns “trying” from missing to entailed in Table~\ref{tab:verification_example_revised_prompt}, improving semantic coverage from 75\% to 100\% and reducing interpretive ambiguity for downstream T2V generation.

In summary, the revision step enables targeted disambiguation by minimally editing underspecified semantics, thereby improving atom-level coverage and reducing interpretive ambiguity for downstream T2V generation.

\subsubsection{Qualitative Comparison of Generated Videos}
\begin{figure*}[ht]
    \centering
    \begin{subfigure}[t]{\textwidth}
        \centering
        \includegraphics[width=\textwidth]{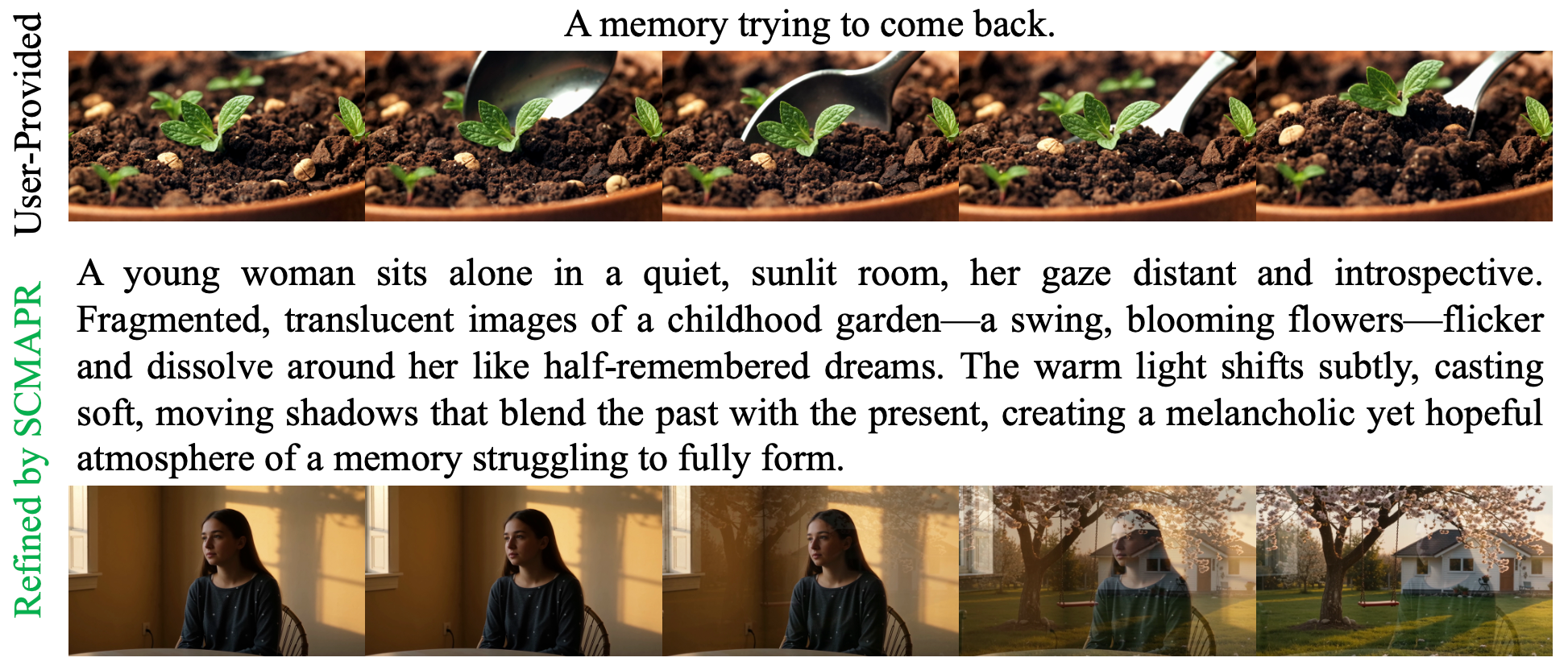}
        \caption{\textbf{Scenario}: Abstract Description. \textbf{Failure Reason (user-provided prompt):} Semantically misaligned with user intention. \textbf{Success Reason (refined prompt):} The refined prompt transfers the abstract notion to concrete elements (a lonely person in a sunlit room) and specifies an explicit recall mechanism, thereby generating a reasonable and user-required video.
        }
    \end{subfigure}
    
    \vspace{2mm}
    
    \begin{subfigure}[t]{\textwidth}
        \centering
        \includegraphics[width=\textwidth]{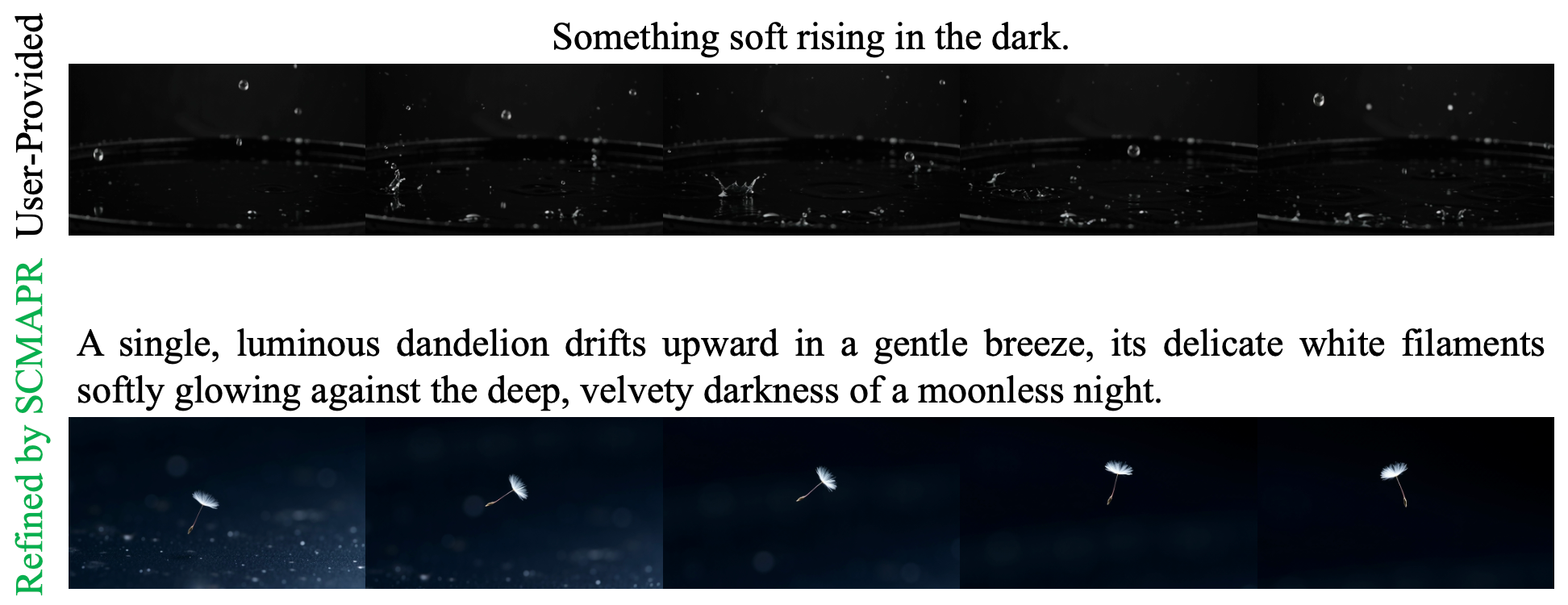}
        \caption{\textbf{Scenario}: Abstract Description. \textbf{Failure Reason (user-provided prompt):} User intends to depict an object rising, while the generated video instead shows raindrops falling. \textbf{Success Reason (refined prompt): } The scene of a dandelion slowly rising aligns with the user-intended depiction.
        }
    \end{subfigure}

    \caption{\textbf{Qualitative comparison of generated videos under a user input (user-provided prompt) and refined prompt from T2V-Complexity benchmark.}
    }
    \label{fig:case_study_video_results_2n3}
\end{figure*}

\begin{figure*}[ht]
    \centering
    \begin{subfigure}[t]{\textwidth}
        \centering
        \includegraphics[width=\textwidth]{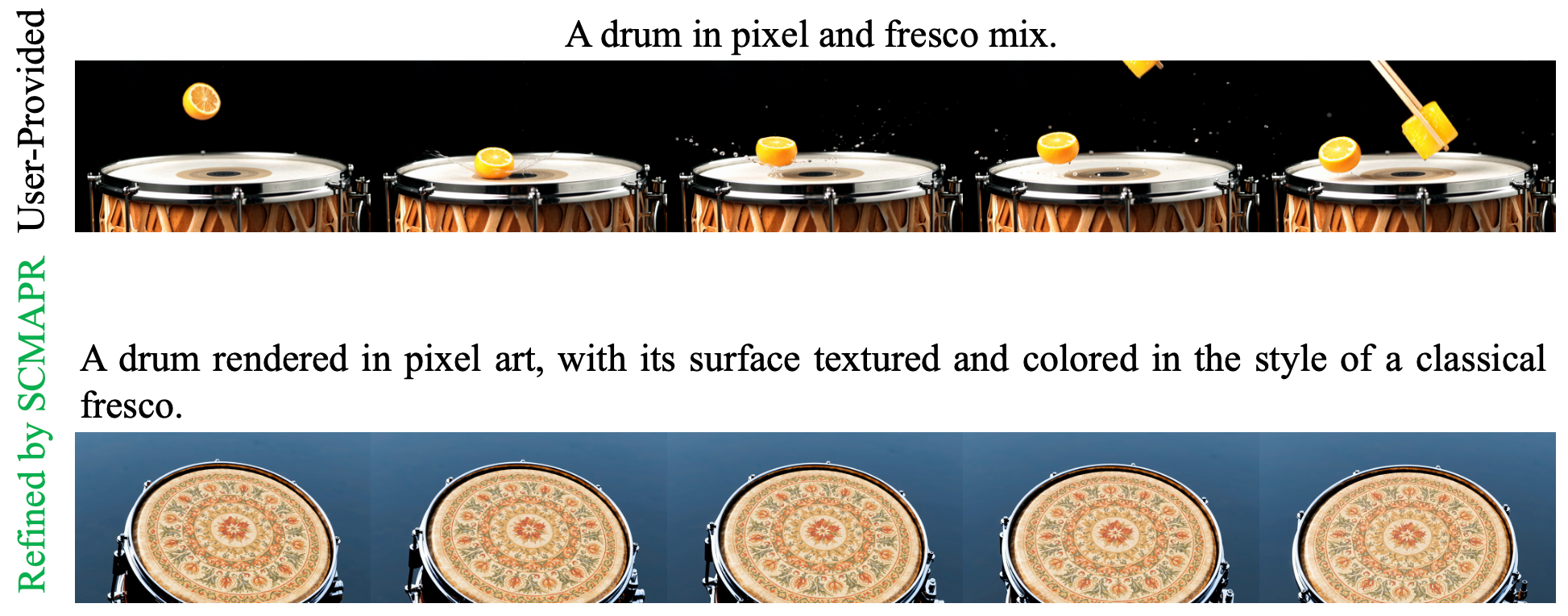}
        \caption{\textbf{ Complex Scenario}: Stylistic Hybrids. \textbf{Failure Reason (user-provided prompt):} The generated video is inconsistent with common sense and misaligned with the user intent. \textbf{Success Reason (refined prompt):} The refined prompt localizes the style mixture to actionable attributes, thereby producing a video with faithful mixed-style.
        }
    \end{subfigure}
    
    \vspace{2mm}
    
    \begin{subfigure}[t]{\textwidth}
        \centering
        \includegraphics[width=\textwidth]{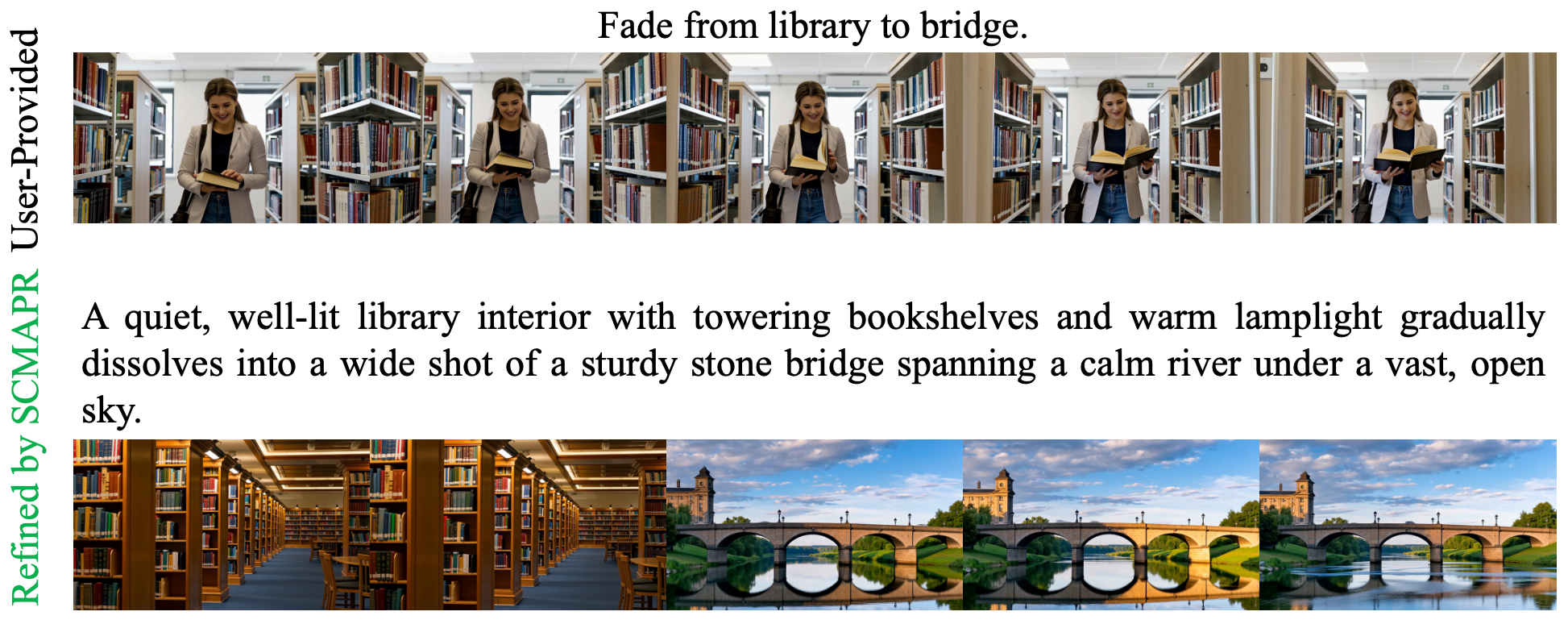}
        \caption{\textbf{Complex Scenario}: Scene Transitions. \textbf{Failure Reason (user-provided prompt):} The video remains confined to the library setting throughout, failing to realize the intended scene transition. \textbf{Success Reason (refined prompt): } The refined prompt explicitly defines both endpoints (library and bridge) and prescribes the transition mechanism (gradual dissolve).
        With such a clear specification, the generated video successfully realizes the scene transition expected by the user.
        }
    \end{subfigure}

    \caption{\textbf{Qualitative comparison of generated videos under a user input (user-provided prompt) and refined prompt from T2V-Complexity benchmark.}
    }
    \label{fig:case_study_video_results_4n5}
\end{figure*}

To visually assess the effect of prompt refinement, Figure~\ref{fig:case_study_video_results} shows sampled frames generated from the user-provided prompt and from the prompt refined by \name{}. With the original abstract prompt, the T2V model drifts to an under-specified dark scene and produces water-ripple imagery that is weakly related to the intended notion of ``hope blooming in the dark.'' In contrast, the refined prompt specifies the abstract intent into concrete elements, namely a flower bud emerging through cracked soil under faint light, together with a warm glow as a symbol of hope. This specification yields output that better preserves the intended subject and atmosphere across frames, resulting in a more coherent and semantically aligned video. Qualitative comparisons of generated videos based on four user inputs and corresponding refined prompts are shown in Figures~\ref{fig:case_study_video_results_2n3} and \ref{fig:case_study_video_results_4n5}.

\clearpage

\end{document}